\documentclass[preprint,times,5p]{elsarticle}
\usepackage{soul}
\usepackage{url}
\usepackage[hidelinks]{hyperref}
\usepackage[utf8]{inputenc}
\usepackage[small]{caption}
\usepackage{graphicx}
\usepackage{amsmath}
\usepackage{amsthm}
\usepackage{booktabs}
\usepackage[noend]{algpseudocode}
\usepackage{algorithmicx,algorithm}
\usepackage{amsfonts}
\usepackage{bm}
\usepackage{subfigure}
\usepackage{cite}
\usepackage{multirow}
\usepackage{longtable}
\usepackage{color}
\usepackage{epstopdf}

\begin{document}
\begin{frontmatter}

\title{Metric-Learning-Assisted Domain Adaptation}

\author{Yueming Yin\corref{cor1}\fnref{fn1}}
\ead{1018010514@njupt.edu.cn}
\author{Zhen Yang\corref{cor1}\fnref{fn2}}
\ead{yangz@njupt.edu.cn}
\author{Haifeng Hu\fnref{fn3}}
\ead{huhf@njupt.edu.cn}
\author{Xiaofu Wu\fnref{fn3}}
\ead{xfuwu@ieee.org}
\fntext[fn1]{School of Telecommunication and Information Engineering, Nanjing University of Posts and Telecommunications, Nanjing 210003, China.}
\fntext[fn2]{Key Lab of Broadband Wireless Communication and Sensor Network Technology, Ministry of Education, Nanjing University of Posts and Telecommunications, Nanjing 210003, China.}
\fntext[fn3]{National Engineering Research Center of Communications and Networking, Nanjing University of Posts and Telecommunications, Nanjing 210003, China.}

\cortext[cor1]{Corresponding author.}

\begin{abstract}
Domain alignment (DA) has been widely used in unsupervised domain adaptation. Many existing DA methods assume that a low source risk, together with the alignment of distributions of source and target, means a low target risk. In this paper, we show that this does not always hold. We thus propose a novel metric-learning-assisted domain adaptation (MLA-DA) method, which employs a novel triplet loss for helping better feature alignment. {\color{blue}{We explore the relationship between the second largest probability of a target sample's prediction and its distance to the decision boundary. Based on the relationship, we propose a novel mechanism to adaptively adjust the margin in the triplet loss according to target predictions.}} Experimental results show that the use of proposed triplet loss can achieve clearly better results. We also demonstrate the performance improvement of MLA-DA on all four standard benchmarks compared with the state-of-the-art unsupervised domain adaptation methods. Furthermore, MLA-DA shows stable performance in robust experiments.
\end{abstract}

\end{frontmatter}
\newpageafter{abstract}

\section{Introduction}
Deep learning approaches have significantly improved a wide variety of machine-learning tasks and computer vision applications. Unfortunately, the impressive performance gains come only when massive amounts of labeled data are available. In practice, manual labeling of such data to train a deep model is often prohibitive or impossible, especially for a target task with no labeled data, e.g. biological images \cite{caicedo2019nucleus}, or a target task with a large number of samples, e.g. video object detection and retrieval \cite{kanazawa2019learning}. Therefore, there is a strong motivation to build the effective learners that can leverage rich labeled data from a different source domain \cite{pan2009survey} or even synthesis data \cite{bashivan2019neural}. However, due to dataset bias or domain shift, predictive models trained on a large-scale dataset do not generalize well to a new dataset or task \cite{pan2009survey}. This learning paradigm suffers from the shift in data distribution across different domains, which poses a huge obstacle for adapting models to the target task \cite{quionero2009dataset}.

\begin{figure}[htbp]
	\centering
	\subfigure[Domain Alignment]{
		\begin{minipage}[c]{0.4\linewidth}
			\centering
			\includegraphics[width=1\linewidth]{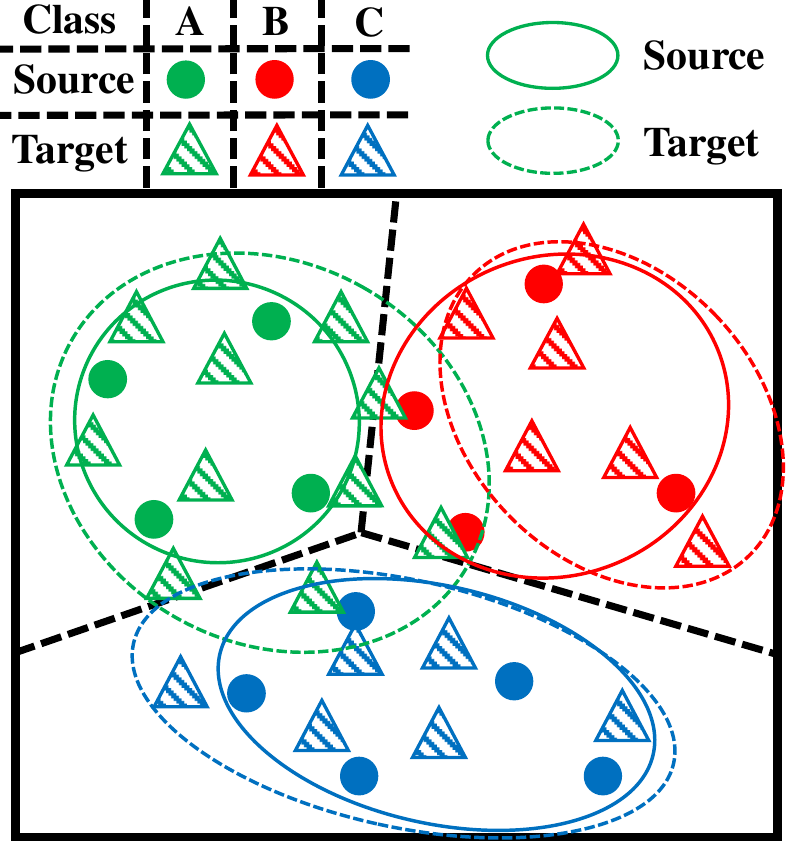}
			\label{insuff_src}
		\end{minipage}}
	\subfigure[Metric-Learning-Assisted]{
		\begin{minipage}[c]{0.4\linewidth}
			\centering
			\includegraphics[width=1\linewidth]{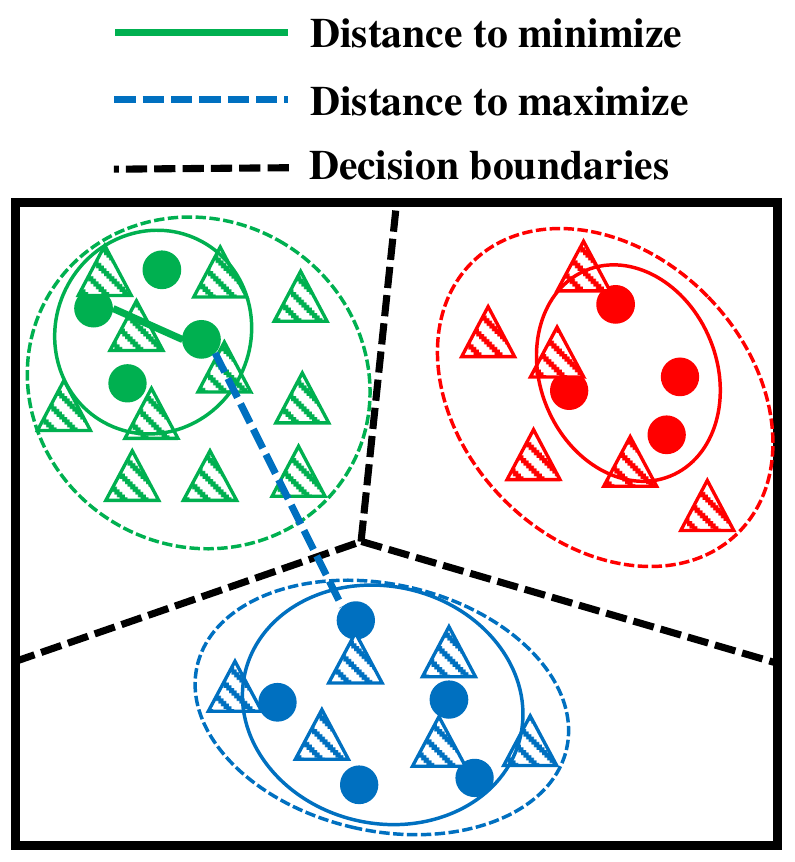}
			\label{insuff_align}
		\end{minipage}}
	\caption{An illustration of (a) domain alignment and (b) metric-learning-assisted domain alignment (MLA-DA). The main idea of MLA-DA is to separate the target features indirectly by separating the source features and aligning the target features with separated source features.}
	\label{WSDA_align}
\end{figure}

Many existing DA methods assume that a low source risk, together with the alignment of distributions of source and target, means a low target risk. However, if the target samples fall outside the support of the source and the embedding function is sufficiently complex, this assumption does not necessarily hold \cite{shu2018dirt}. Unfortunately, deep learning models are complex enough to be overfitting, especially higher layer neurons are more sensitive to the original task, but not suitable for the target task \cite{yosinski2014transferable}. The limitation of DA only is illustrated in Fig. \ref{WSDA_align}. We observe that target samples are distributed around the decision boundary, or even misaligned to the other side when aligned to the source domain. Hence, one of the main goals of the proposed MLA-DA is to separate these target features from the decision boundaries.

\begin{figure*}[htbp]
	\centering
	\subfigure[Prediction]{
		\begin{minipage}[c]{0.3\textwidth}
			\centering
			\includegraphics[width=1\linewidth]{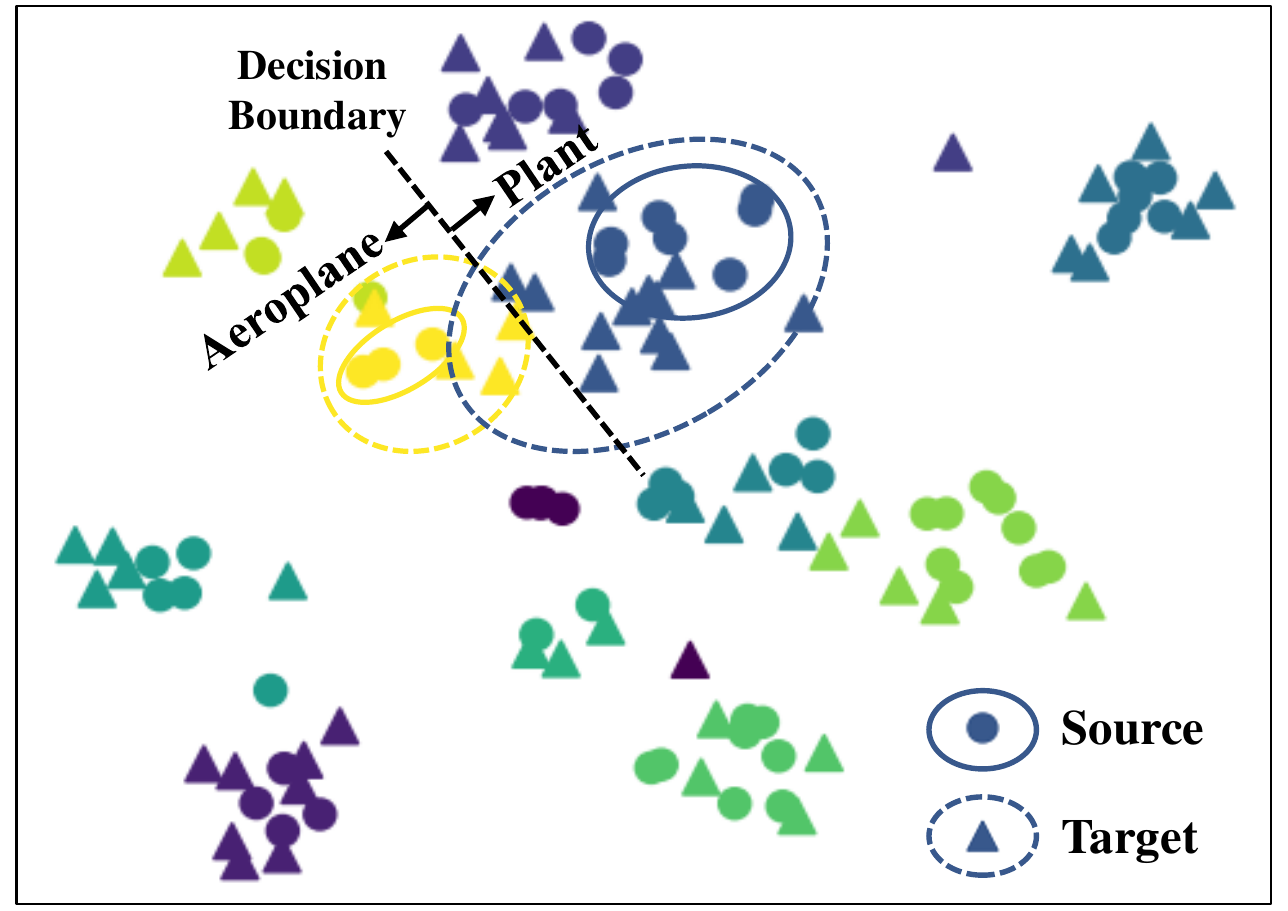}
			\label{pseudo}
		\end{minipage}}
	\subfigure[True Label]{
		\begin{minipage}[c]{0.3\textwidth}
			\centering
			\includegraphics[width=1\linewidth]{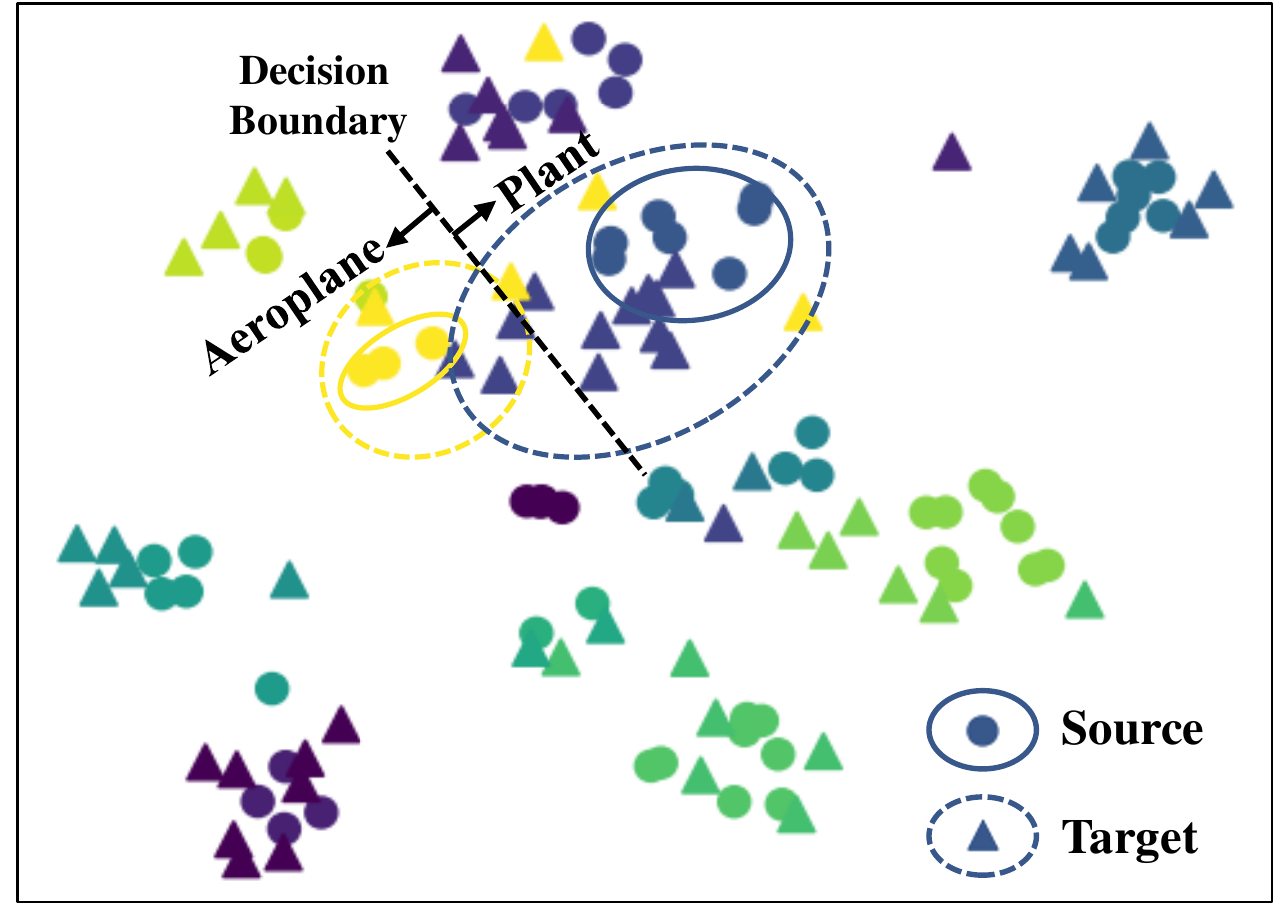}
			\label{true}
		\end{minipage}}
	\subfigure[{\color{blue}{Prediction probabilities of a target sample near the decision boundary}}]{
		\begin{minipage}[c]{0.3\textwidth}
			\centering
			\includegraphics[width=1\linewidth]{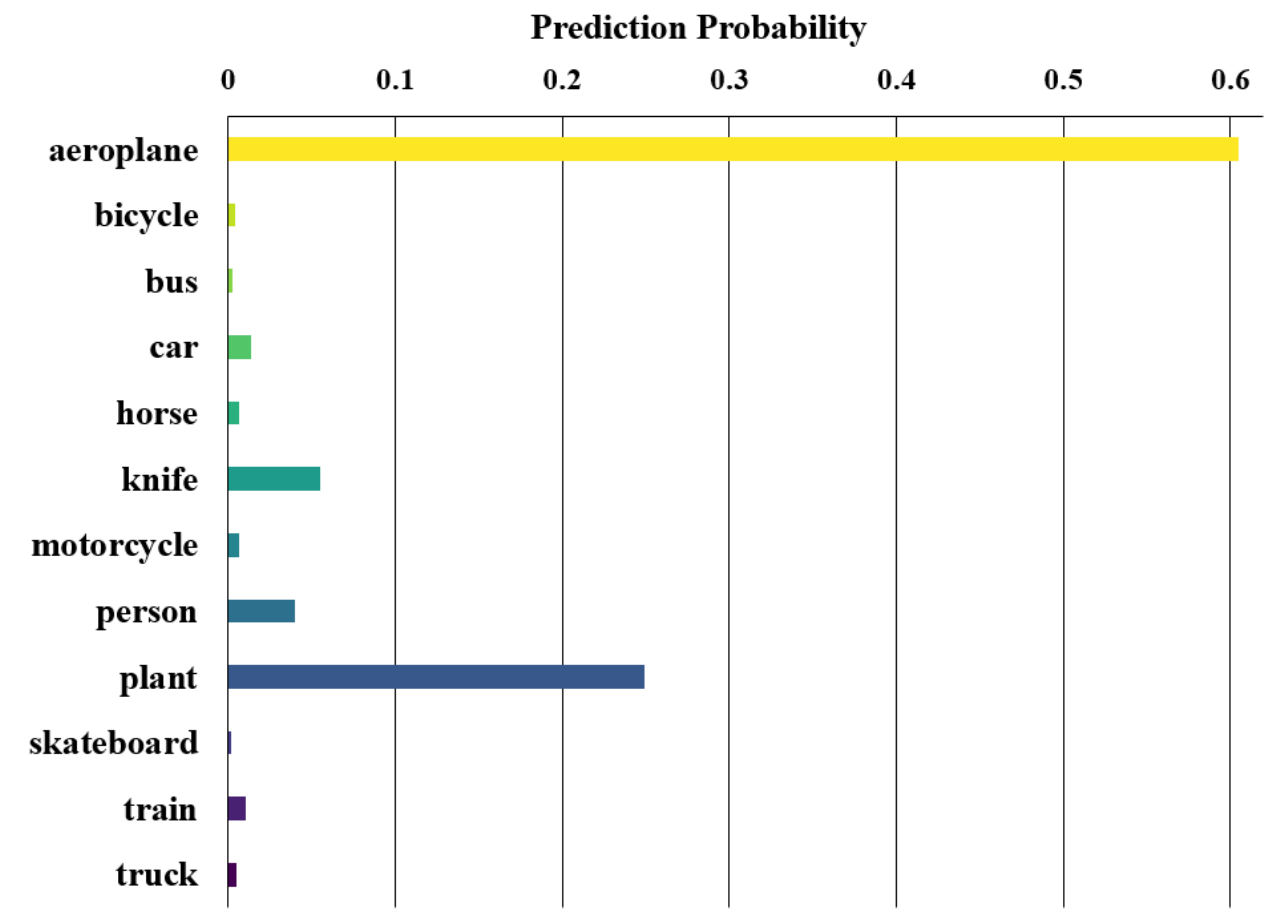}
			\label{subpeak}
		\end{minipage}}
	\caption{\textbf{An example for the discovery of misclassified samples}. (a) and (b) are t-SNE visualization on VisDA dataset trained only by the source label loss and the domain alignment loss. Samples in figure (a) are colored with predicted labels, and the same samples are colored with their true labels in figure (b). (c) is the predicted probability of target samples belonging to every class, which are predicted as the aeroplane class as shown in (a) and (b). The  relationships between colors and labels are shown in figure (c).}
	\label{fsubpeak}
\end{figure*}

Intuitively, the closer input sample gets to the decision boundaries, the more uncertain the corresponding classifier's output. Therefore, the probabilities that a target sample belongs to different categories are used to identify the desired target sample. Specifically, when the prediction is not confident, the probability of prediction as a wrong category is greater as shown in Fig. \ref{subpeak}. Moreover, these two categories are close in the feature space as shown in Fig. \ref{pseudo}. So we use the second largest value of the prediction probability to measure how close a sample is to the decision boundary.

The next task is how to push these target samples away from the decision boundaries. Since the target sample has no label, this is in general difficult. Fortunately, domain alignment can generally align the feature distribution of source and target domains. So we can indirectly push target samples away from the decision boundaries by increasing large enough margin for different classes in source domain. Specifically, when the feature distributions are aligned, the interval between neighboring source categories is broaden to separate target samples as many as possible. Hence, the misclassification rate of target samples can be reduced further.

To do this task, we design a new triplet loss to learn a better feature extractor. The main idea of proposed triplet loss includes two points: features of positive pairs (samples with the same label) should be indistinguishable and features of negative pairs (samples with the different labels) should be discriminative both in the source and target domains. However, there is loss of the target discriminative information during domain alignment. {\color{blue}{So, we firstly adaptively the margin in the triplet loss according to target predictions. Based on the above analysis, we add the second largest probability of target prediction to the margin of the mostly likely label. For any target sample near the decision boundary, the use of the second largest prediction probability as the margin will push the learned features near the decision boundary to be more discriminative.}} The major contributions of this work can be summarized as follows: 
\begin{itemize}
    {\color{blue}{\item We explore the relationship between the second largest probability of a target sample's prediction and its distance to the decision boundary. Based on the relationship, we propose a novel mechanism to adaptively adjust the margin in the triplet loss according to target predictions.}}
    \item We proposed a metric-learning-assisted domain adaptation (MLA-DA) to push target samples away from the decision boundaries by applying a triplet loss with dynamic margin. To the best of our knowledge, this is the first work to adopt the margin of triplet loss in unsupervised domain adaptation, which achieves clearly better results.
    \item Extensive experimental results on four standard benchmarks demonstrate that proposed MLA-DA achieves superior performance compared with state-of-the-art unsupervised domain adaptation methods. And robust experimental results demonstrate that MLA-DA has stable performance even if the source domain size is reduced.
\end{itemize}

\section{Related Work}
\subsection{Domain Alignment}
Learning a discriminative classifier or other predictors in the presence of the shift between training and test distributions is known as transfer learning or domain adaptation \cite{pan2009survey}. The main technical difficulty of previous domain adaptation is how to formally reduce the distribution discrepancy across different domains. To address this issue, a variety of domain adaptation approaches have been proposed \cite{long2015learning,ganin2016domain,tzeng2017adversarial,saito2018maximum,zhang2019bridging}. Recently, numerous adversarial adaptation methods \cite{ajakan2014domain,ganin2016domain,kim2017learning,tzeng2017adversarial,ma2019gcan} have been proposed, which borrow the essential idea from generative adversarial network (GAN) \cite{goodfellow2014generative}. In these adversarial domain adaptation methods, a domain classifier is trained to tell whether the sample comes from the source domain or target domain. Meanwhile, the feature extractor is trained to minimize the classification loss and maximize the domain confused loss. Discriminative and domain-invariant features can be obtained through adversarial training.

{\color{blue}{Recently, many impressive adversarial domain adaptation methods \cite{DBLP:conf/cvpr/ZhangTJT19,DBLP:conf/cvpr/ChangYSKH19,DBLP:conf/cvpr/ChenXHRD0XH19,ma2019gcan} have been proposed. A novel domain-symmetric networks was proposed in \cite{DBLP:conf/cvpr/ZhangTJT19} based on a symmetric design of source and target task classifiers. Unlike other adversarial domain adaptation methods, \cite{DBLP:conf/cvpr/ZhangTJT19} designed a specific domain confusion loss for the feature extractor to maximally confuse the two domains, instead of confusing the domain discriminator. A domain-specific batch normalization method was proposed by \cite{DBLP:conf/cvpr/ChangYSKH19}, adopting specific batch normalization strategies for both domains. In \cite{DBLP:conf/cvpr/ChenXHRD0XH19}, a progressive feature alignment method has been proposed to align the discriminative features across domains progressively, via exploiting the intra-class variation in the target domain. With the invention of graph convolutional network \cite{DBLP:conf/iclr/KipfW17}, a novel graph convolutional adversarial network was proposed by \cite{ma2019gcan}, jointly modeling data structure, domain label, and class label in a unified deep model.}}

As shown in Fig. \ref{pseudo}, feature distribution of source and target samples are aligned through domain adversarial training. When the overall feature distribution extracted from the source and target domains cannot be distinguished by the discriminator, the extracted features are considered domain-invariant. These domain-invariant features maintain the discriminative information of source domain and the decision boundaries trained from source data are also used to distinguish target samples. 

\subsection{Triplet Loss in Metric Learning}
Metric learning concerns learning a reasonable metric over the input space, and it has attracted considerable attention recently \cite{weinberger2009distance,xing2003distance,davis2007information,zuo2017distance,cheng2017duplex}. Xing et al. learned a good distance metric for similar point pairs by respecting these relationships \cite{xing2003distance}. Weinberger et al. presented a Mahalanobis distance function for the k-nearest neighbors (kNN) classifier by utilizing a triplet loss that forces exemplars from the same class to be clustered together, while exemplars from different classes are effectively separated \cite{weinberger2009distance}. Davis et al. proposed an information-theoretic Mahalanobis distance metric approach by minimizing the differential relative entropy between two distance functions \cite{davis2007information}.

In \cite{radenovic2016cnn} and \cite{simo2015discriminative}, a siamese model was trained with a pairwise loss in deep metric learning. One of the most studied pairwise losses is the contrastive loss \cite{chopra2005learning}, which minimizes the distance between positive pairs and maximizes the distance between negative pairs as long as this “negative distance” is smaller than a margin.

The triplet loss  is proposed in \cite{schroff2015facenet,harwood2017smart,qian2015fine} to handle the issue that the optimization of the positive pairs is independent from the negative pairs, but the optimization should force the distance between positive pairs to be smaller than negative pairs in pairwise loss. It is defined based on three samples: an anchor sample, a positive sample (i.e., a sample belonging to the same class as the anchor), and a negative sample (i.e., a sample from a different class of the anchor). The loss will force the positive pair distance plus a margin to be smaller than the negative pair distance.

\subsection{Entropy Minimization}
Entropy minimization (EM) was first proposed in \cite{grandvalet2005semi} for semi-supervised learning. It was argued in \cite{morerio2018minimal} that EM could be achieved by the optimal alignment of second order statistics between source and target domains and therefore a hyper-parameter validation method was proposed for balancing the reduction of the domain shift and the supervised classification on the source domain in an optimal way. In \cite{cariucci2017autodial}, a novel domain alignment layer was introduced for reducing the domain shift by aligning source and target distributions to a reference one and entropy minimization was also explicitly employed, which was believed to promote classification models with high confidence on unlabeled samples. \cite{long2015learning} used EM in their approach to directly measure how far samples are from a decision boundary by calculating entropy of the classifier's output. In the appendix of \cite{saito2018maximum}, which proposed an entropy-based adversarial dropout regularization approach to employ the entropy of target samples in implementing min-max adversarial training. In \cite{long2018conditional}, entropy conditioning was employed that controlled the uncertainty of classifier predictions to guarantee transferability, which can help the proposed Conditional Adversarial Domain Adaptation (CDAN) to converge to better solutions.

\section{Metric-Learning-Assisted Domain Adaptation}
In this section, we provide details of proposed MLA-DA.

\subsection{Preliminaries}
In the scenario of the unsupervised domain adaptation, we define $n_s$ labeled samples $\left\{ \left(  \bm{x}_{s}^{\left( i \right)},y_{s}^{\left( i \right)} \right) \right\}_{i=1}^{n_s}$ from the source joint distribution $\mathcal{D}_s$, where $\bm{x}_{s}^{(i)} \in \mathcal{X}_{S}$ and $y_{s}^{(i)} \in \mathcal{Y}_{S}$. $\mathcal{X}_{S}$ and  $\mathcal{Y}_{S}$ denote the source data space and source label space, respectively. Similarly, we also define $n_t$ unlabeled target samples $\left\{\left(\bm{x}_{t}^{(i)}\right)\right\}_{i=1}^{n_{t}}$, where $\bm{x}_{t}^{(i)} \in \mathcal{X}_{T}$, and the $\mathcal{X}_{T}$ represents target data space drawn from the target joint distribution $\mathcal{D}_t$. The $\mathcal{X}_{S}$ and $\mathcal{X}_{T}$ are assumed to be different but related (referred as covariate shift in \cite{shimodaira2000improving}). The goal of unsupervised domain adaptation is to develop an embedding function $F: \{\mathcal{X}_S,\mathcal{X}_T\} \rightarrow \mathbb{R}^n$ and a classifier $C: \mathbb{R}^n \rightarrow \mathbb{R}^k$, such that the classifier $H=C\circ F$ is able to predict the labels for samples from the target domain. The domain classifier $D: \mathbb{R}^n \rightarrow \{0,1\}$ predicts the probability of a sample $x$ belonging to source ($D(x)=1$) or target domain ($D(x)=0$). Moreover, we introduce a metric generator $G: \mathbb{R}^n \rightarrow \mathbb{R}^m$. The number of classes is $k$, i.e. the source label set $\mathcal{Y_S}=\{0,1,\cdots,k\}$.

\subsection{Limitations of Domain Alignment}
Though target samples are aligned to the source samples, some target samples might still be near the decision boundary as shown in Fig. \ref{pseudo} and Fig. \ref{true}, and three samples of plants fall into the area of airplanes after alignment.

For the joint distribution of source domain $\mathcal{D}_s$, we define the risk of the classifier $H=C\circ F$ by:
\begin{equation}
\epsilon _{\mathcal{D}_s}(h)=\mathbb{E}_{(\bm{x},y)\in \mathcal{D}_s}\mathbf{1}\{y\neq\arg\max _{y_i}\hat{P}(y_i|\bm{x},H)\},
\label{source risk}
\end{equation}where $\mathbf{1}\{\cdot\}=1$ if $\{\cdot\}$ is true. $\arg\max _{y_i}\hat{P}(y_i|x,H)$ is the probability of $x$ belonging to the i-th class predicted by classifier $H$. DA aims to learn a single classifier $H$ used for both source and target domains. Therefore, domain adversarial training of DA sets up the objective:
\begin{align}
&\min _{F \in \mathcal{F}, C \in \mathcal{C}}\epsilon _{\mathcal{D}_s}(C\circ F)
\nonumber\\ & s.t.\ F(\mathcal{X}_S)=F(\mathcal{X}_T),
\label{DA objective}
\end{align}where $\mathcal{F}$ and  $\mathcal{C}$ are the hypothesis space for the embedding function and embedding classifier. Meanwhile, a domain classifier $D: \mathbb{R}^n \rightarrow \{0,1\}$ is trained in DA to satisfy the constraint in Eq. \ref{DA objective} by:
\begin{align}
&\max _{F \in \mathcal{F}}\mathbb{E}_{\bm{x_s}\in \mathcal{X}_S}\mathbf{1}\{D(\bm{x})\neq 1\}+\mathbb{E}_{\bm{x_t}\in \mathcal{X}_T}\mathbf{1}\{D(\bm{x})\neq 0\}
\nonumber\\ & \min _{D \in \mathcal{D}}\mathbb{E}_{\bm{x_s}\in \mathcal{X}_S}\mathbf{1}\{D(\bm{x})\neq 1\}+\mathbb{E}_{\bm{x_t}\in \mathcal{X}_T}\mathbf{1}\{D(\bm{x})\neq 0\},
\label{domain classifier}
\end{align}where $\mathcal{D}$ is the hypothesis space for the domain classifier $D$. When the trained discriminator is still unable to distinguish the source and target features, the extracted features are shown in the Fig. \ref{pseudo}. The source and target feature distributions are similar, but compared to Fig. \ref{true}, we can find that some samples spread to the other side of the decision boundary due to misalignment. This misalignment often occurs between two similar categories, and it is difficult to correct the misclassified target samples caused by misalignment.

\subsection{MLA-DA Loss}
\label{DMA}
In order to avoid aligned samples falling into other categories, one can resort to metric learning for possible way out. In this paper, we introduce a new triplet loss:
\begin{figure*}[htbp]
	\centering
	\includegraphics[width=0.8\linewidth]{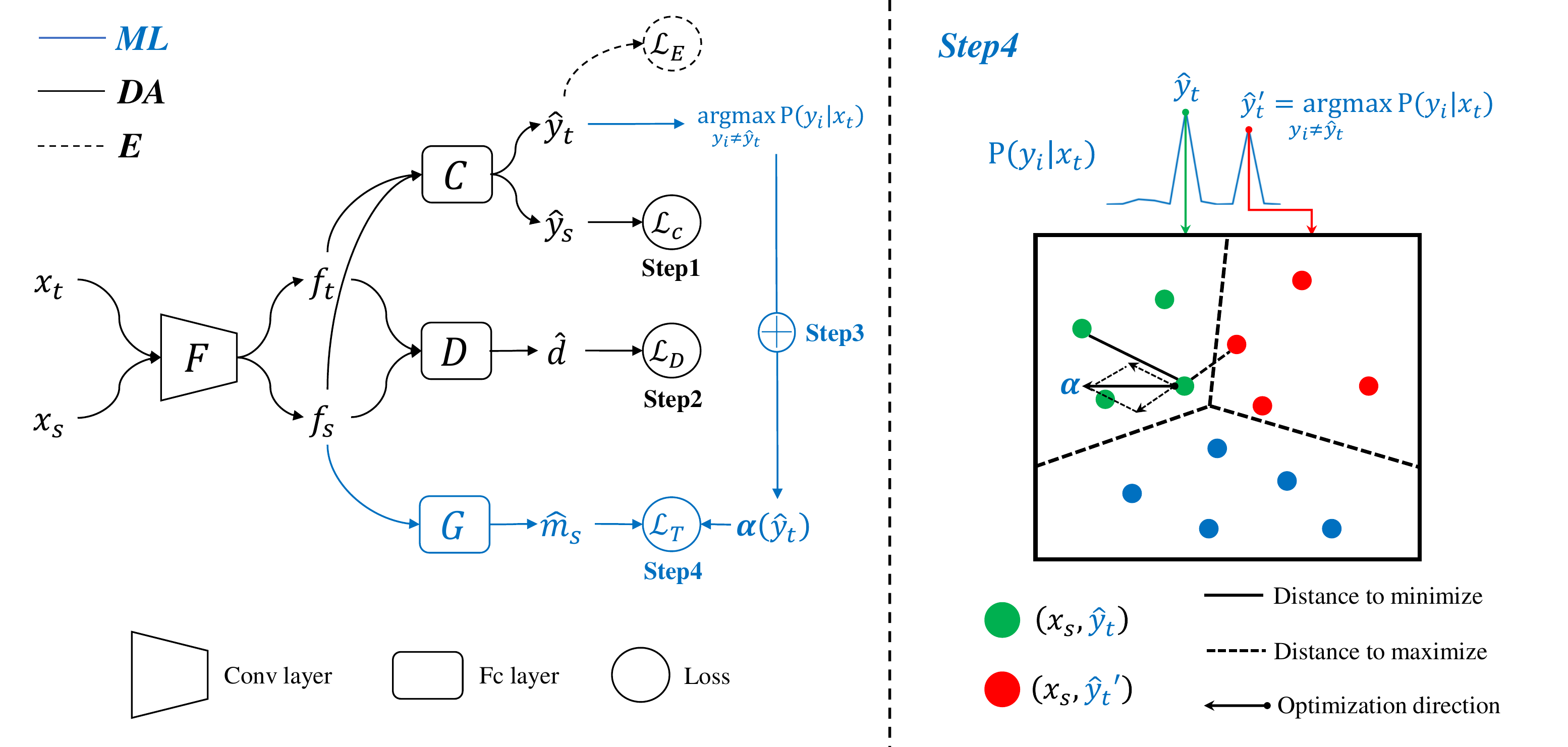}
	\caption{\textbf{Metric-Learning-Assisted Domain Adaptation (MLA-DA)}. The MLA-DA networks consists of four steps in every iteration. \textbf{Step1}: Fine-tune Feature Extractor $F$ and update Classifier $C$ for samples from source distribution $(x_s,y_s)$. \textbf{Step2} (\textbf{D}omain \textbf{A}lignment): Align features extracted from source and target data distribution $f_s$ and $f_t$ by adversarial Discriminator $D$. \textbf{Step3}: Update margin $\alpha$ with the probability of the second possible label and minimize entropy \textbf{(E)} loss for target samples. \textbf{Step4} (\textbf{M}etric \textbf{L}earning): Minimize triplet loss with updated $\alpha$ for source metric feature $\hat{m}_s$. The blue part is our contribution. The right part illustrates the step4 in detail.}
	\label{Total_Model}
\end{figure*}

\begin{align}
\mathcal{L}_{T}{\color{blue}{(\mathcal{X}_S,\mathcal{Y}_S,\mathcal{X}_T)}}&=\frac{1}{b} \sum_{i=1}^{b} \max (\max _{y_i=y_{j}}\|M(\bm{x_i})-M(\bm{x_{j}})\|^{2}
\nonumber\\&-\min _{y_i \neq y_{k}}\|M(\bm{x_i})-M(\bm{x_{k}})\|^{2}+\alpha (y_i), 0),
\nonumber\\& (\bm{x_i}, y_i),(\bm{x_j}, y_j),(\bm{x_k}, y_k) \in D^{batch}_s,
\label{Triplet_Loss}
\end{align}where $M:\mathcal{X} \rightarrow \mathbb{R}^m$ is the metric function learning by embedding function $F: \mathcal{X} \rightarrow \mathbb{R}^n$ and metric generator $G: \mathbb{R}^n \rightarrow \mathbb{R}^m$, $b$ is the batch size, $D^{batch}_s$ is a batch of samples drawn from source distribution.{\color{blue}{ Input target distribution $\mathcal{X}_T$ is used to calculate the dynamic margin $\alpha (y_i)$ shown in Eq. \ref{alpha}.}} In Eq. \ref{Triplet_Loss}, the first term is the maximum distance between positive pairs $(\bm{x_i},\bm{x_j})$ to decrease the intra-class distance of source features, the second term is the minimum distance between negative pairs $(\bm{x_i},\bm{x_k})$ to increases the inter-class distance, and the third term $\alpha (y_i)$ is the adjustable margin for class $y_{i}$ in the metric space. This loss increases the discrimination interval of different classes and reduce the interval of same classes simultaneously

Generally, the closer the sample is to the decision boundary, the more uncertain the corresponding classifier output is. Therefore, the probabilities that the target sample belongs to different categories are used to judge whether this sample is close to the corresponding decision boundary. As shown in Fig. \ref{subpeak}, except for the aeroplane with the highest probability, we also consider the plant with the second highest probability. {\color{blue}{Since the input pictures are resized to a same size, some categories are difficult to be distinguished by the classifier.}} Especially, the sample whose second highest probability is comparatively higher is close to the decision boundary and can be easily misclassified as shown in Fig. \ref{pseudo}. So we use the second largest value of the prediction probability $\max _{y\neq \hat{y}} \hat{P}(y|x)$ to measure how close a sample is to the decision boundary{\color{blue}{. For target samples near the decision boundary, the use of $\max _{y\neq \hat{y}} \hat{P}(y|x)$ in the margin might encourage them to move away from decision boundaries. For target samples far from decision boundaries, the use of $\max _{y\neq \hat{y}} \hat{P}(y|x)$ in the margin also makes sense since their values are often approaching zeros. Therefore,}}  $\alpha (\hat{y}_i)$ is defined as follows in the same mini-batch:
\begin{equation}
\alpha (\hat{y}_i)=\alpha_{0}+\mu \frac{1}{b} \sum_{i=1}^{b}\max _{y\neq \hat{y}_i} \hat{P}(y|\bm{x_i}),\bm{x_i} \in \mathcal{X}_{T}^{batch},
\label{alpha}
\end{equation}where $\hat{P}(y|\bm{x})$ is the probability of $x$ belonging to the class $y$ predicted by the classifier $H=G\circ F$. $\alpha_{0}$ is the initial value, and $\mu$ is the constant coefficient. $\hat{y}_i=\mathop{\arg\max}_y \hat{P}(y|\bm{x})$ is the pseudo label of the target sample $\bm{x_i}\in \mathcal{X}_{T}^{batch}$. The pseudo label can be obtained by a classifier trained in advance from source data, and these obtained probabilities contain the discriminative information of target domain. {\color{blue}{In Eq. \ref{alpha}, target samples are divided into groups according to their pseudo labels, and the margin of each label is calculated by averaging $\max _{y\neq \hat{y}} \hat{P}(y|x)$ over the corresponding group of target samples.}}

A larger margin in triplet loss is introduced to push these easily misclassified classes further away from each other, and simultaneously force samples in one class to cluster together in the feature space. This ensure that different categories are separated by a large enough margin. Moreover, the proposed triplet loss is computed in a mini-batch during training, which adjusts source label distribution efficiently and avoids run-time complexity exploding mentioned in \cite{do2019theoretically}.

\subsection{Target Separation by MLA-DA}
The task of domain adaptation is to obtain a robust transfer classifier that performs well on the target domain, and the result of the classification depends on the embedding function $F: \{\mathcal{X}_S,\mathcal{X}_T\} \rightarrow \mathbb{R}^n$. We hope that the features of the target samples learned by $F$ are easy to classify: the samples of the same category are as close as possible, and the samples of different categories are as far away as possible. However, in the unsupervised domain adaptation, the label of the target sample is unknown, and we cannot directly keep the target sample away from decision boundaries. Instead, we alternately force the source features away from the decision boundary and align the feature distributions of the source and target domains.

The architecture of MLA-DA is shown in Fig. \ref{Total_Model}. It consists of a feature extractor $F$, an adversarial domain discriminator $D$, a metric generator $G$ and a label classifier $C$. Input $x$ from either domain is fed into the feature extractor $F$. The extracted features $f_s=F(x_s)$, $f_t=F(x_t)$ are forwarded into the label classifier $C$ to obtain the softmax output $\hat{y_s}=C(f_s)$, $\hat{y_t}=C(f_t)$ over all classes. The adversarial domain discriminator $D$ aims to adversarial match the feature distribution of the source and target data. The Metric Generator $G$ obtains metric feature $\hat{m}_s=G(f_s)$ from source feature. The total loss of MLA-DA is:
\begin{align}
\mathcal{L}_{Total}&=\mathcal{L}_{C}\left(\mathcal{X}_{S},\mathcal{Y}_{S}\right) + \mathcal{L}_{D}\left(\mathcal{X}_{S}, \mathcal{X}_{t}\right)
\nonumber\\ &+\gamma \mathcal{L}_{T}\left(\mathcal{X}_{S},\mathcal{Y}_{S},\mathcal{X}_{T}\right) + \lambda \mathcal{L}_{E}\left(\mathcal{X}_{T}\right).
\label{Total_Loss_E}
\end{align}

The total loss is computed and optimized in every batch as shown in Algorithm \ref{alg:MAN}. The classification loss $\mathcal{L}_{C}$ is shown in Eq. \ref{Class_Loss}:
\begin{equation}
\mathcal{L}_{C}=\mathbb{E}_{(\bm{x}, y) \in D_{S}}[L(C(F(\bm{x})), y)],
\label{Class_Loss}
\end{equation}the $L(\cdot, \cdot)$ is typically a cross entropy loss for supervised classification. According to \cite{goodfellow2014generative}, the domain alignment loss is shown in Eq. \ref{Adv_Loss}. Specifically, we employ a domain classifier $D$ as discriminator to tell whether the feature embeddings from feature extractor $F$ arise from source or target data distribution, while the $F$ is trained to fool $D$ by a gradient reversal layer between $F$ and $D$. This two-player minimax game is expected to reach an equilibrium where the feature embeddings from $F$ are domain-invariant.
\begin{align}
\mathcal{L}_{D}=&-\mathbb{E}_{\bm{x} \in \mathcal{X}_{S}} \log D(F(\bm{x})) 
\nonumber\\&-\mathbb{E}_{\bm{x} \in \mathcal{X}_{T}} \log (1-D(F(\bm{x}))).
\label{Adv_Loss}
\end{align}

Followed \cite{grandvalet2005semi}, we implement target entropy minimization in MLA-DA to enforce the decision boundaries pass through low-density area in the target domain. The target entropy loss is shown in Eq. \ref{L_E}:
\begin{equation}
\mathcal{L}_{E}=-\frac{1}{\left|\mathcal{X}_{t}\right|} \sum_{\bm{x} \in \mathcal{X}_{t}}\mathcal{H}(C(F(\bm{x}))),
\label{L_E}
\end{equation}the $\mathcal{H}(\cdot)$ is the entropy function. Due to label loss $\mathcal{L}_{C}$ and domain alignment loss $\mathcal{L}_{D}$ are both computed with cross-entropy loss function of source label and domain label, we give the same weight for them. The $\gamma$ and $\lambda$ are the balance parameters for triplet loss $\mathcal{L}_{T}$ and target entropy loss $\mathcal{L}_{E}$, respectively. The algorithm of proposed MLA-DA is shown as Algorithm \ref{alg:MAN}.

\begin{algorithm}[htb]
\caption{MLA-DA Algorithm}
\label{alg:MAN}
{\bf Input:}\\
$\mathcal{X}_S=\{ \textbf{x}_{s}^{\left( i \right)}\}_{i=1}^{n_s}$: source training sample set;\\
$\mathcal{Y}_S=\{ y_{s}^{\left( i \right)}\}_{i=1}^{n_s}$: source training label set;\\
$\mathcal{X}_T=\{ \textbf{x}_{t}^{\left( i \right)}\}_{i=1}^{n_t}$: target training sample set;\\
$F: \mathcal{X} \rightarrow \mathbb{R}^n$: embedding function parameterized by $\theta _F$;\\
$C: \mathbb{R}^n \rightarrow \mathbb{R}^k$: embedding classifier parameterized by $\theta _C$;\\
$D: \mathbb{R}^n \rightarrow [0,1]$: domain classifier parameterized by $\theta _D$;\\
$G: \mathbb{R}^n \rightarrow \mathbb{R}^m$: metric generator parameterized by $\theta _G$;\\
$T$: max iteration.
\begin{algorithmic}[1]
\State load parameters pre-trained on ImageNet for $\theta _F$;
\State set $t=0$;
\For{each batch $(\mathcal{X}^{batch}_{S}, \mathcal{Y}^{batch}_{S}, \mathcal{X}^{batch}_{T})$ in $(\mathcal{X}_S, \mathcal{Y}_S, \mathcal{X}_T)$}
	\State calculate $\mathcal{L}_{C}$ for $\mathcal{D}^{batch}_{S}$ by Eq. \ref{Class_Loss};
	\State obtain target prediction $\hat{P}(y_i|x)$ for $x\in \mathcal{X}^{batch}_{T}$ by $f$;
	\State calculate $\mathcal{L}_{D}$ for $(\mathcal{X}^{batch}_{S}, \mathcal{X}^{batch}_{T})$ by Eq. \ref{Adv_Loss};
	\State calculate $\mathcal{L}_{E}$ for $\mathcal{X}^{batch}_{T}$ by Eq. \ref{L_E};
	\State calculate $\alpha (y_i)$ for $\mathcal{X}^{batch}_{T}$ by Eq. \ref{alpha};
	\State calculate $\mathcal{L}_{T}$ for $\mathcal{D}^{batch}_{S}$ by Eq. \ref{Triplet_Loss};
	\State calculate total loss $\mathcal{L}_{Total}$ by Eq. \ref{Total_Loss_E};
	\State update $\theta_F $, $\theta _C $, $\theta _D$ and $\theta _G$ to minimize $\mathcal{L}_{Total}$ by statistical gradient descent;
	\State let $t\leftarrow t+1$;
	\If{$t=T$}
		\State \textbf{break.}
	\EndIf
\EndFor
\end{algorithmic}
{\bf Output:}
$\theta_F $, $\theta _C $, $\theta _D$ and $\theta _G$.
\end{algorithm}

\section{Experiments}
In this section, we present extensive experimental results and analyze the robustness of proposed MLA-DA.
\begin{figure*}[htbp]
	\centering
	\subfigure[$\mathcal{L}_{C}$]{
		\begin{minipage}[c]{0.25\textwidth}
			\centering
			\includegraphics[width=0.8\linewidth]{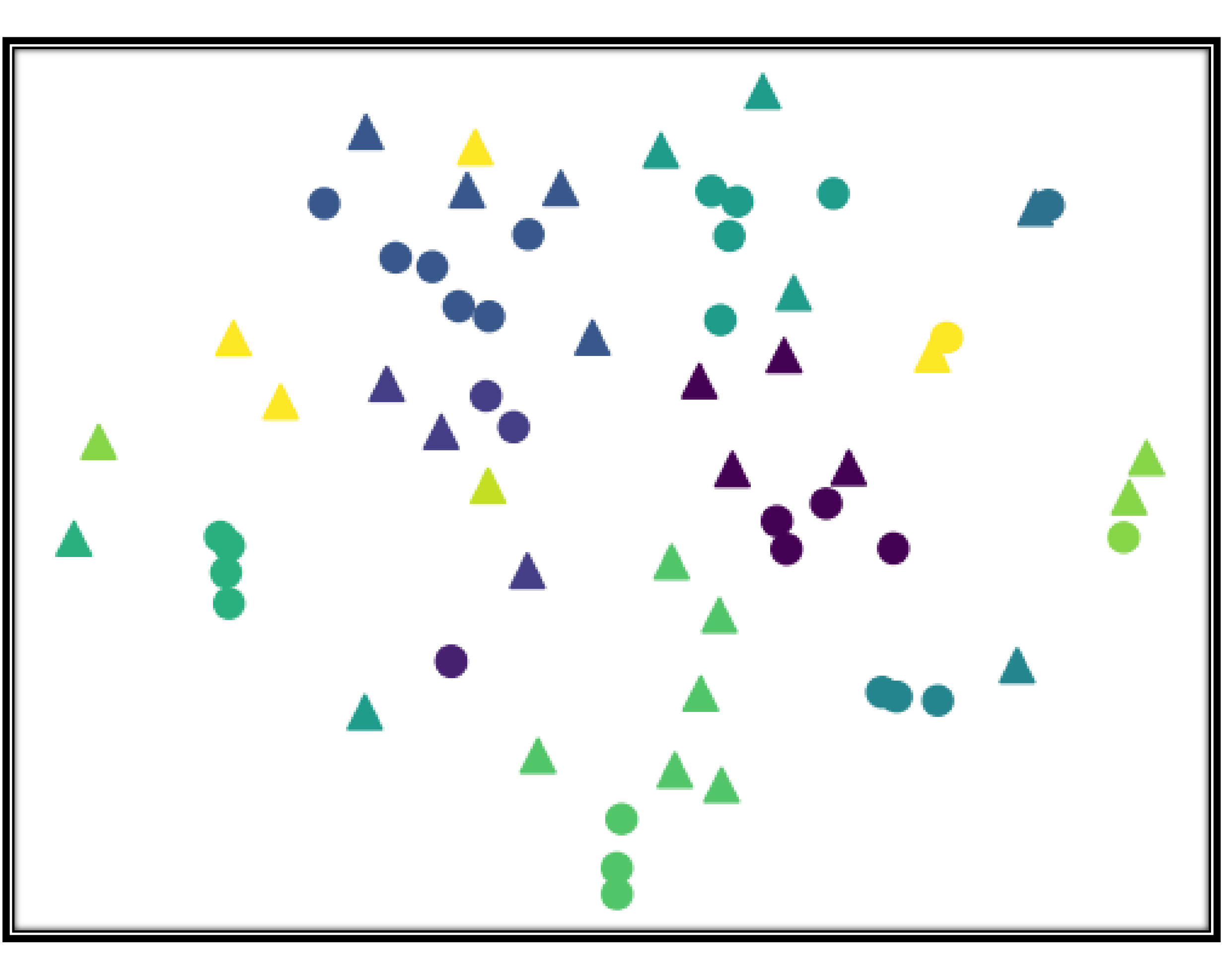}
			\label{src}
		\end{minipage}}
	\subfigure[$\mathcal{L}_{C}+\mathcal{L}_{D}$]{
		\begin{minipage}[c]{0.25\textwidth}
			\centering
			\includegraphics[width=0.8\linewidth]{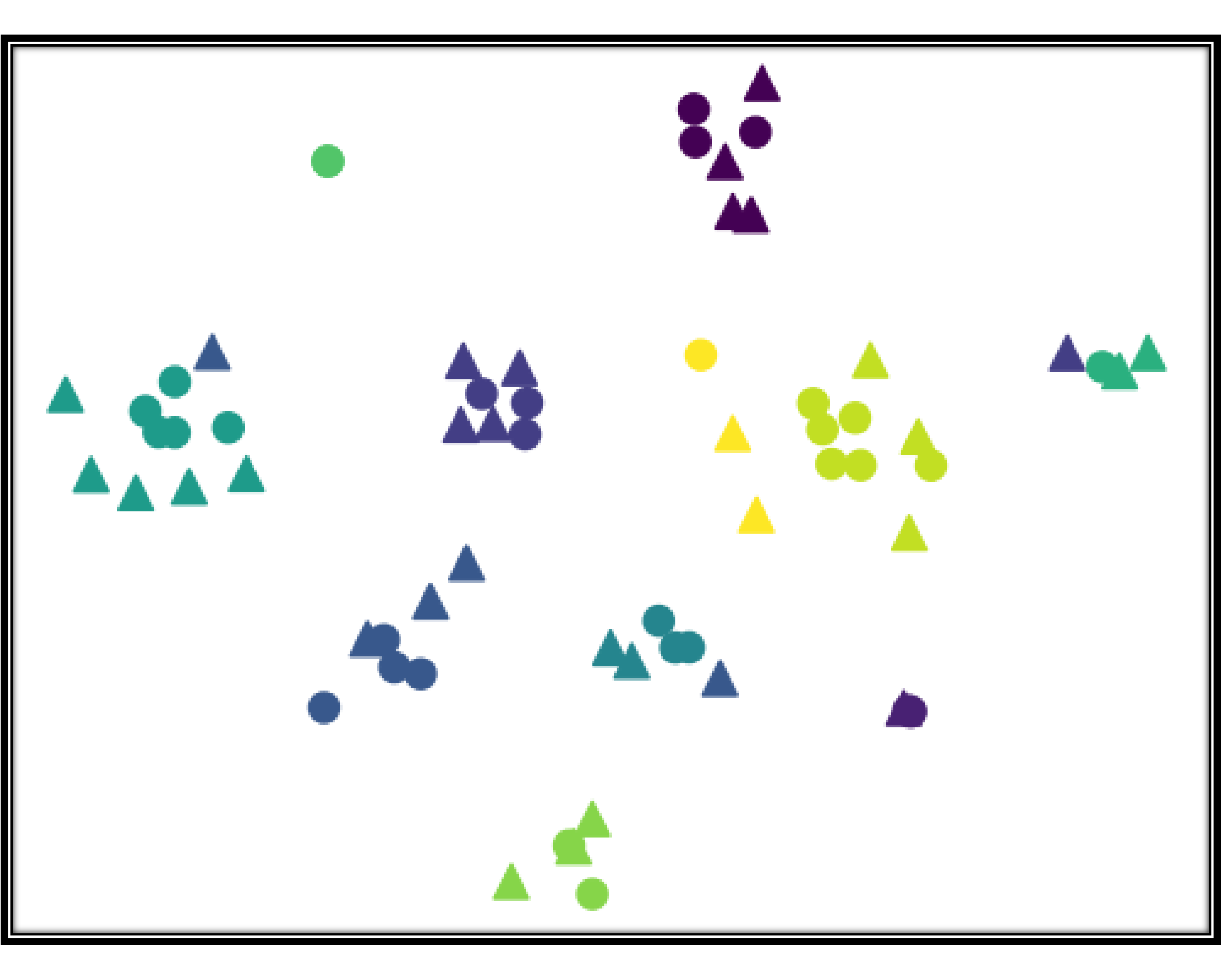}
			\label{adv}
		\end{minipage}}
	\subfigure[$\mathcal{L}_{C}+\mathcal{L}_{D}+\gamma \mathcal{L}_{T}$]{
		\begin{minipage}[c]{0.25\textwidth}
			\centering
			\includegraphics[width=0.8\linewidth]{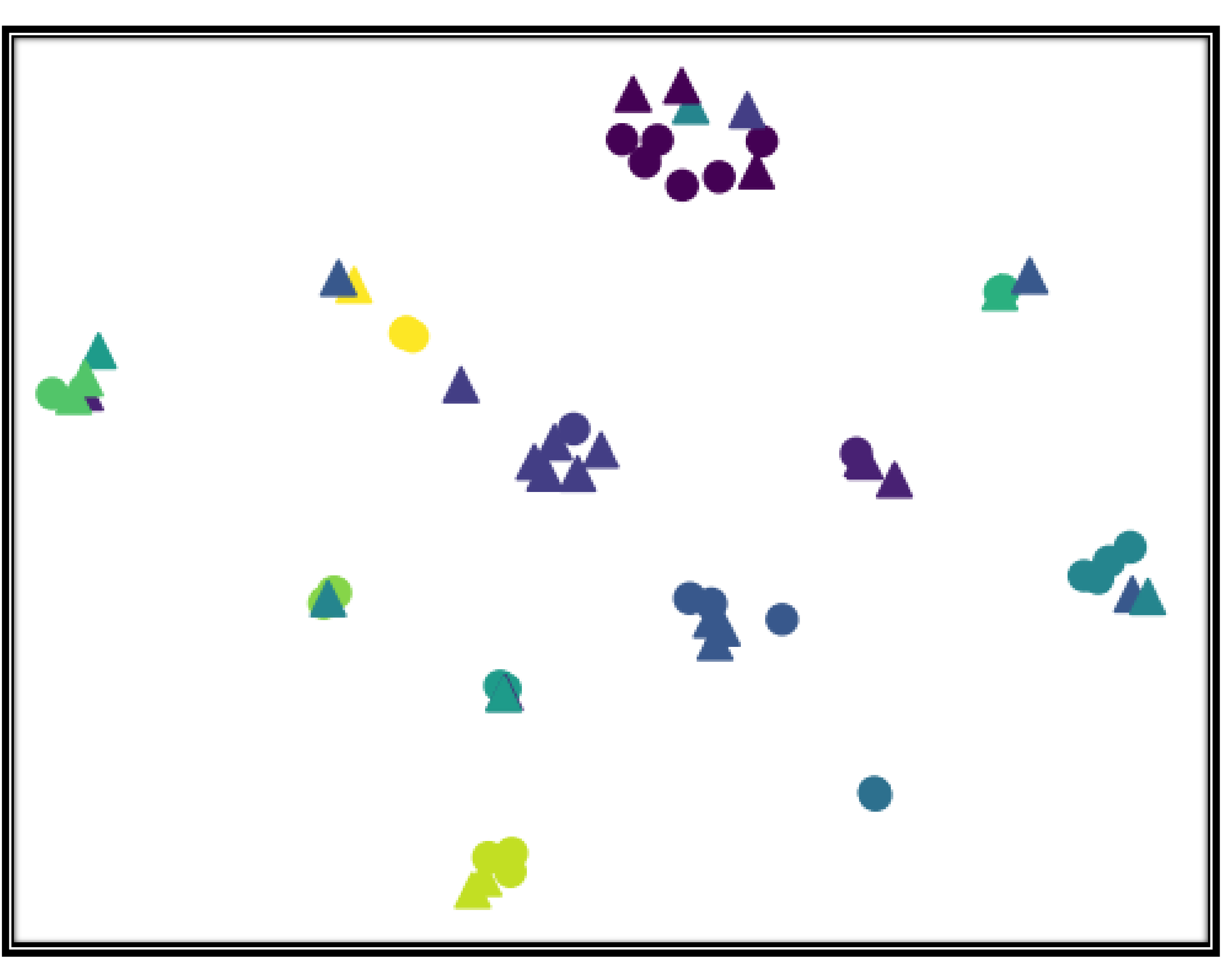}
			\label{metric}
		\end{minipage}}
	\caption{\textbf{t-SNE visualization on VisDA dataset with different loss function}. The circular and triangle point represent the feature embedding of samples from source domain and target domain, respectively. Points are colored with their true labels both in source and target domain.}
	\label{t-SNE}
\end{figure*}

\subsection{Datasets}
\textbf{Office-31} \cite{saenko2010adapting} is a benchmark dataset for domain adaptation, comprising 4,110 images in 31 classes collected from three distinct domains: Amazon (A), which contains images downloaded from amazon.com, Webcam (W) and DSLR (D), which contain images taken by web camera and digital SLR camera with different photographic settings, respectively. To enable unbiased evaluation, we evaluate all methods on all six transfer tasks A$\rightarrow$W, D$\rightarrow$W, W$\rightarrow$D, A$\rightarrow$D, D$\rightarrow$A and W$\rightarrow$A.

\textbf{Office-Home} \cite{venkateswara2017deep} contains 4 domains, each with 65 categories including daily objects. Specifically, Art (Ar) denotes artistic depictions for object images, Clipart (Cl) means picture collection of clipart, Product (Pr) shows object images with a clear background and is similar to Amazon category in Office-31, and Real-World (Rw) represents object images collected with a regular camera. We use all domain combinations and build 12 transfer tasks.

\textbf{VisDA2017} \cite{DBLP:journals/corr/abs-1710-06924} is simulation-to-real dataset with two domains: Synthetic renderings of 3D models generated from different angles and with different lighting conditions and Real collected from photo-realistic or real-image datasets. Since the 3D models were generated in clean environment, the Synthetic domain is very different from Real domain. With 280K images across 12 classes, the scale of VisDA2017 also brings challenges to domain adaptation.

\textbf{ImageCLEF-DA} \footnote{https://www.imageclef.org/2014/adaptation} is a benchmark dataset for ImageCLEF 2014 domain adaptation challenge, which is organized by selecting the common categories shared by the following three public datasets. Here, each dataset is considered as a domain: Caltrch-256 (C), ImageNet ILSVRC 2012 (I), and Pascal VOC 2012 (P). There are 50 images in each category and 600 images in each domain. We consider six transfer tasks: I$\rightarrow$P, P$\rightarrow$I, I$\rightarrow$C, C$\rightarrow$I, C$\rightarrow$P and P$\rightarrow$C.

\begin{table}[h]
    \centering 
    \caption{Accuracy (\%) on D$\rightarrow$A and W$\rightarrow$A task with different weight $\gamma$.} 
    \begin{tabular}{cccc} 
        \toprule 
        $\gamma$ & D$\rightarrow$A & W$\rightarrow$A & Avg \\ 
        \midrule 
        0.01 & 73.1 & 67.8 & 70.5 \\
        0.05 & 73.4 & 70.0 & 71.7 \\
        0.08 & \textbf{74.7} & \textbf{71.0} & \textbf{72.9} \\
        0.1 & 74.0 & 70.5 & 72.3 \\
        0.2 & 73.0 & 70.1 & 71.6 \\
        0.5 & 71.3 & 69.2 & 70.3 \\
	  \bottomrule 
    \end{tabular} 
    \label{gamma} 
\end{table}

\begin{table}[h]
    \centering 
    \caption{Accuracy (\%) on D$\rightarrow$A and W$\rightarrow$A task with different margin $\alpha$.} 
    \begin{tabular}{cccc} 
        \toprule 
        $\alpha$ & D$\rightarrow$A & W$\rightarrow$A & Avg \\ 
        \midrule 
        1 & 73.1 & 68.7 & 70.9 \\
        5 & 73.3 & 69.2 & 71.3 \\
        10 & 74.1 & 70.8 & 72.5 \\
        20 & 74.3 & 70.9 & 72.6 \\
        30 & 73.6 & 68.2 & 70.9 \\
        40 & 73.6 & 70.6 & 72.1 \\
        Eq. \ref{alpha} & \textbf{74.7} & \textbf{71.0} & \textbf{72.9} \\
	  \bottomrule 
    \end{tabular} 
    \label{dynamic} 
\end{table}

\subsection{Baseline Methods}
We compare our \textbf{MLA-DA} with state-of-the-art domain adaptation methods: Deep Adaptation Network (\textbf{DAN}) \cite{long2015learning},Reverse Gradient (\textbf{RevGrad}) \cite{ganin2015unsupervised}, Domain Adversarial Neural Network (\textbf{DANN}) \cite{ganin2016domain}, Joint Adaptation Net (\textbf{JAN}) \cite{long2017deep}, Adversarial Discriminative Domain Adaptation (\textbf{ADDA}) \cite{tzeng2017adversarial}, Multi-Adversarial Domain Adaptation (\textbf{MADA}) \cite{pei2018multi}, Maximum Classifier Discrepancy (\textbf{MCD}) \cite{saito2018maximum}, and Conditional Domain Adversarial Network (\textbf{CDAN}) \cite{long2018conditional}.

\begin{figure*}[h]
	\centering
	\subfigure[Accuracy D$\rightarrow$A]{
		\begin{minipage}[c]{0.24\textwidth}
			\centering
			\includegraphics[width=1\linewidth]{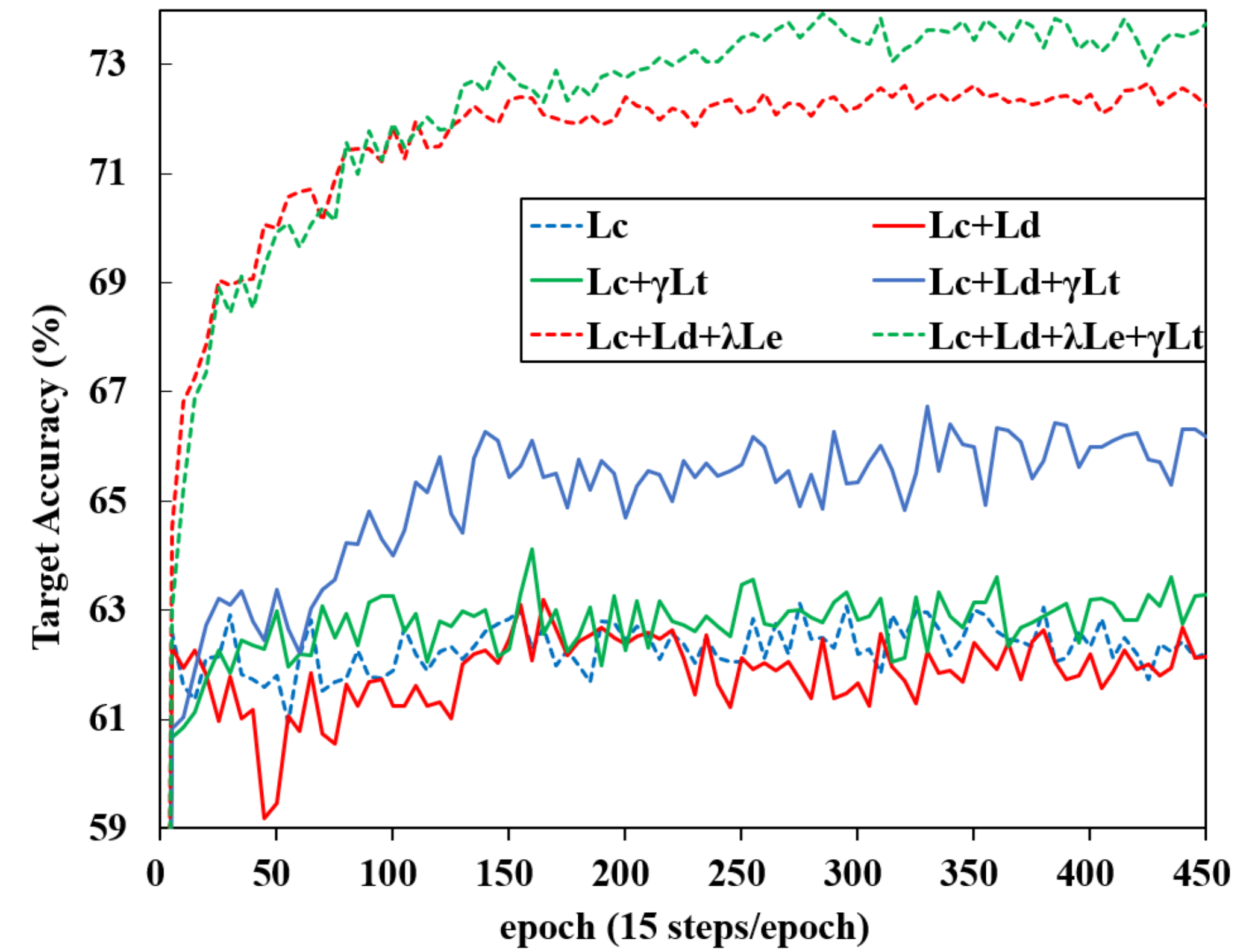}
			\label{D-A_acc}
		\end{minipage}}
	\subfigure[$\mathcal{L}_{T}$ D$\rightarrow$A]{
		\begin{minipage}[c]{0.24\textwidth}
			\centering
			\includegraphics[width=1\linewidth]{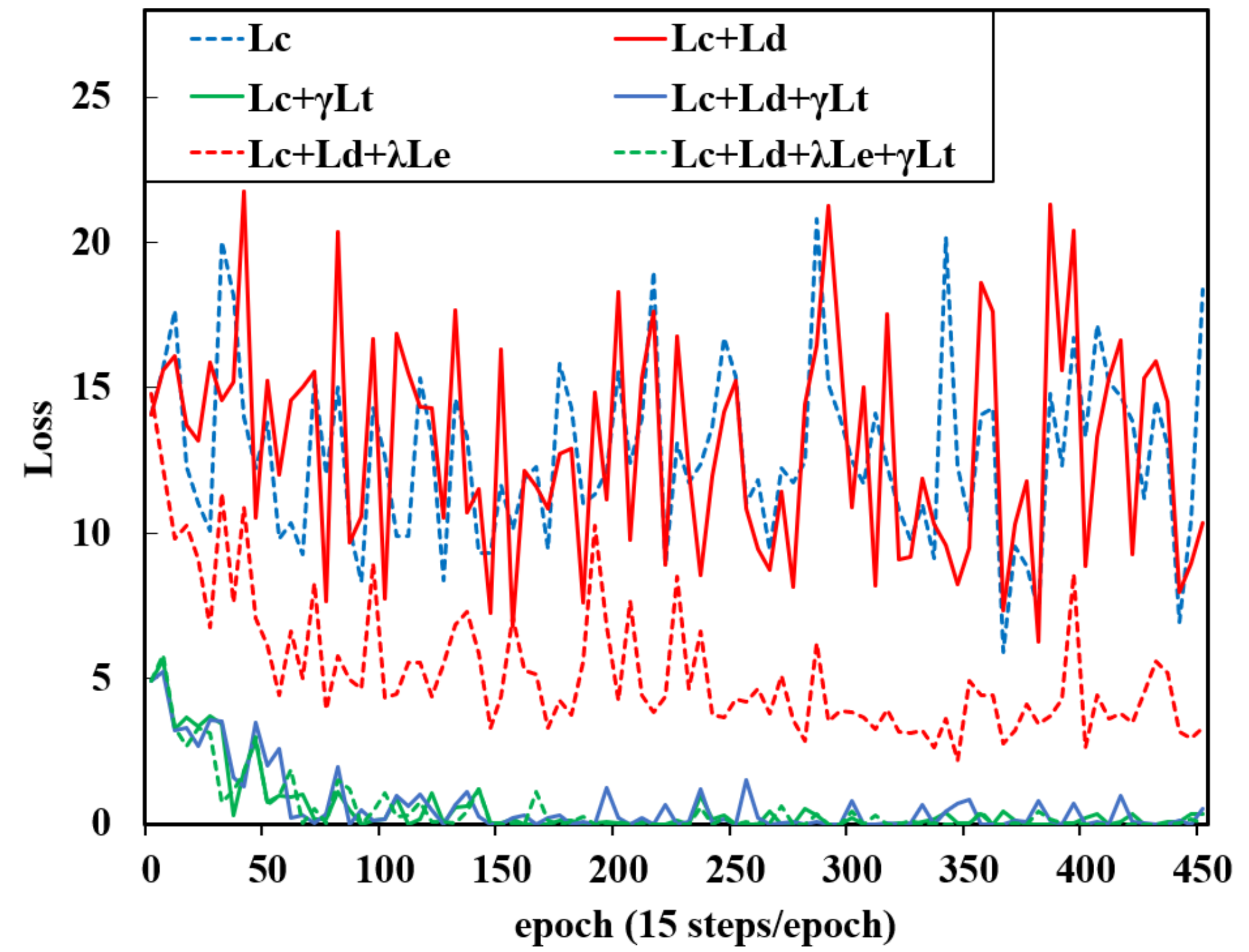}
			\label{D-A_loss}
		\end{minipage}}
	\subfigure[Accuracy W$\rightarrow$A]{
		\begin{minipage}[c]{0.24\textwidth}
			\centering
			\includegraphics[width=1\linewidth]{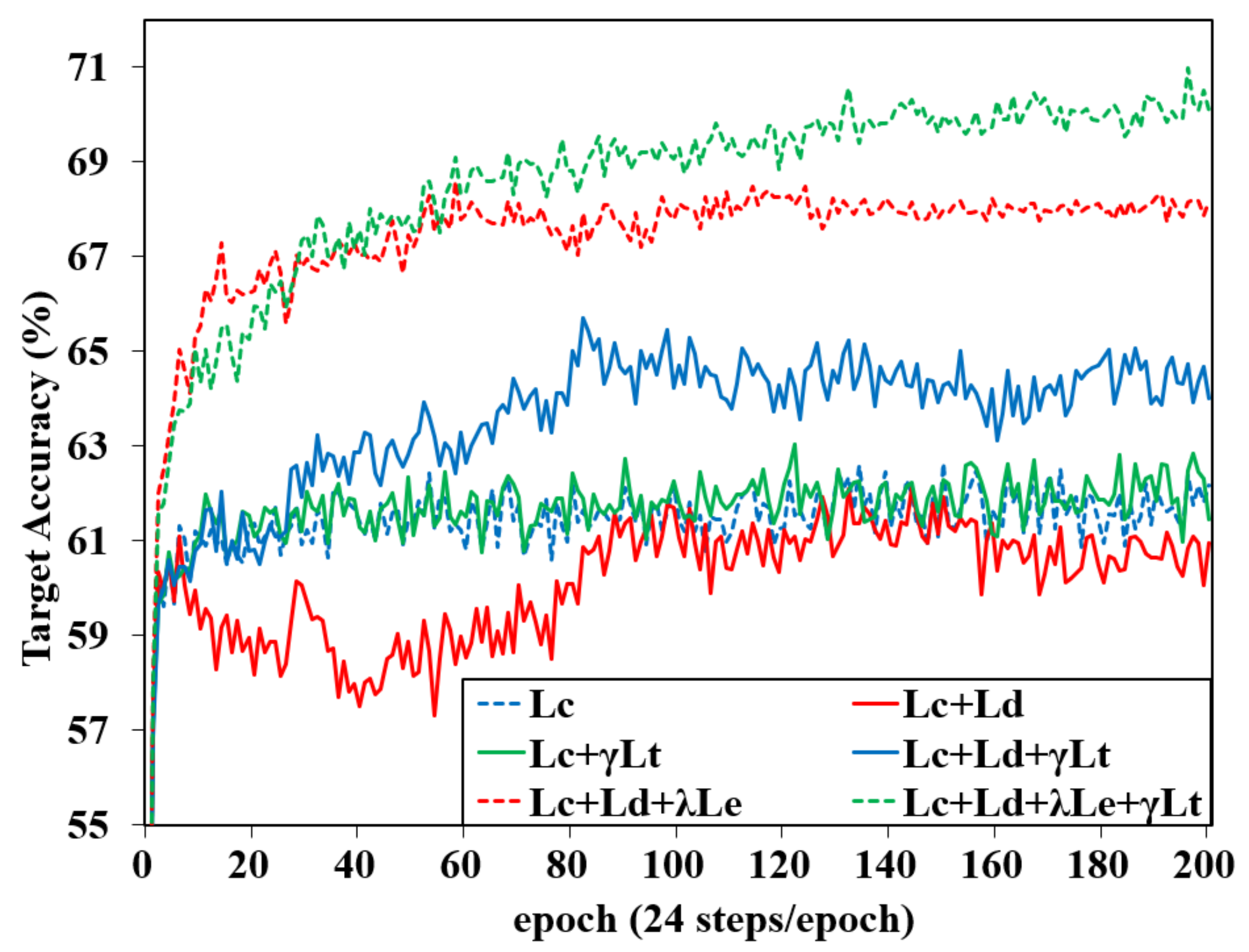}
			\label{W-A_acc}
		\end{minipage}}
	\subfigure[$\mathcal{L}_{T}$ W$\rightarrow$A]{
		\begin{minipage}[c]{0.24\textwidth}
			\centering
			\includegraphics[width=1\linewidth]{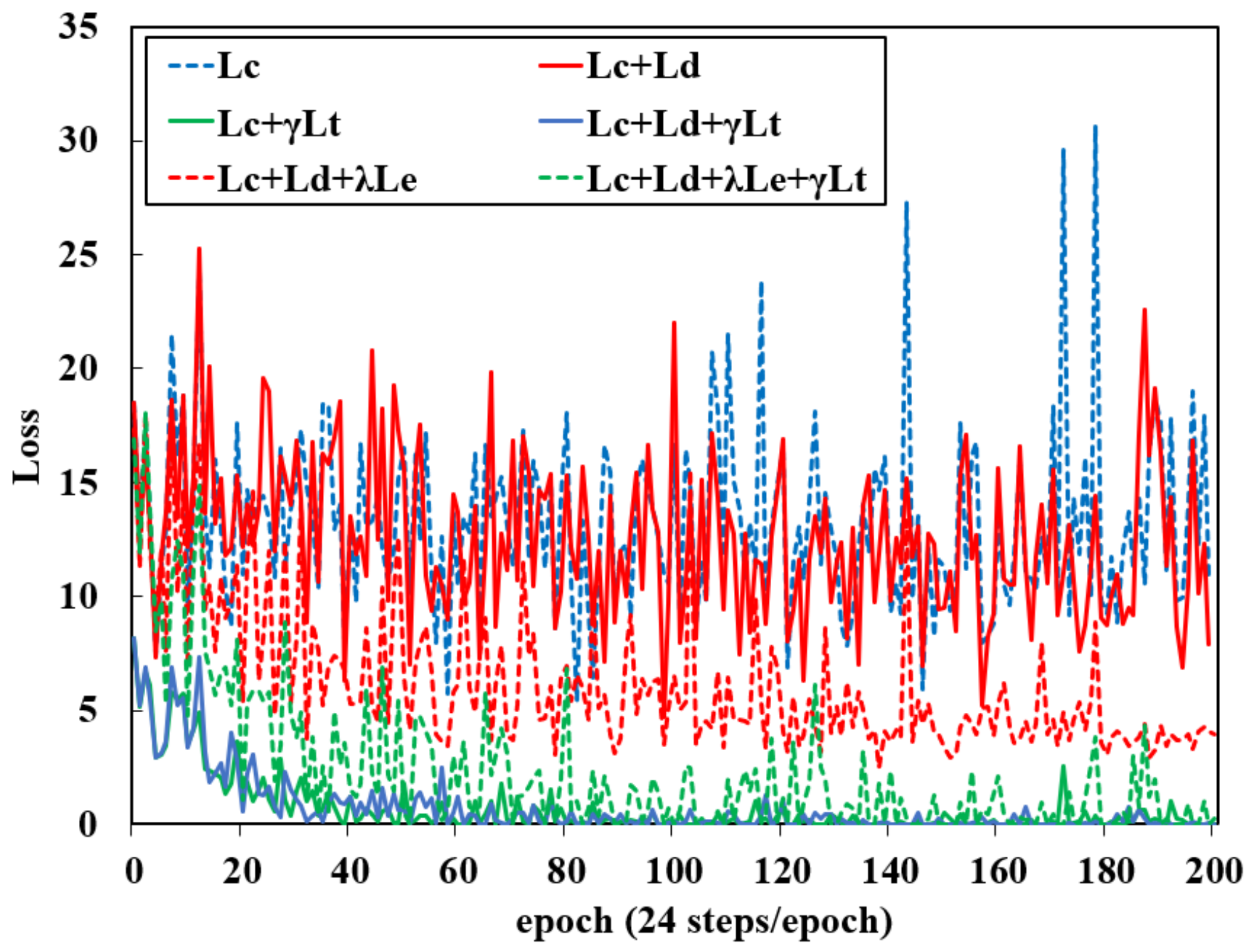}
			\label{W-A_loss}
		\end{minipage}}
	\caption{(a) and (c): Comparison of different loss functions on the target accuracy. (b) and (d): Comparison of different loss functions on the triplet loss $\mathcal{L}_{T}$. Our triplet loss is effective and universal when the domain alignment loss does not work well on the small-to-large transfer tasks D$\rightarrow$A and W$\rightarrow$A.}
	\label{acc_loss}
\end{figure*}

\begin{table*}[h]
    \centering 
    \caption{Accuracy (\%) on Office-31 for unsupervised domain adaptation (ResNet-50)} 
    \begin{tabular}{cccccccc} 
        \toprule 
        Method & A$\rightarrow$W & D$\rightarrow$W & W$\rightarrow$D & A$\rightarrow$D & D$\rightarrow$A & W$\rightarrow$A & Avg \\ 
        \midrule 
        Resnet-50 \cite{he2016deep}& 68.4$\pm$0.2 & 96.7$\pm$0.1 & 99.3$\pm$0.1 & 68.9$\pm$0.2 & 62.5$\pm$0.3 & 62.7$\pm$0.3 & 76.2\\ 
        DAN \cite{long2015learning}& 80.5$\pm$0.4 & 97.1$\pm$0.2 & 99.6$\pm$0.1 & 78.6$\pm$0.2 & 63.6$\pm$0.3 & 62.8$\pm$0.2 & 80.4\\ 
        DANN \cite{ganin2016domain}& 82.0$\pm$0.4 & 96.9$\pm$0.2 & 99.1$\pm$0.1 & 79.7$\pm$0.4 & 68.2$\pm$0.4 & 67.4$\pm$0.5 & 82.2\\ 
        ADDA \cite{tzeng2017adversarial}& 86.2$\pm$0.5 & 96.2$\pm$0.3 & 98.4$\pm$0.3 & 77.8$\pm$0.3 & 69.5$\pm$0.4 & 68.9$\pm$0.5 & 82.9\\ 
        JAN \cite{long2017deep}& 85.4$\pm$0.3 & 97.4$\pm$0.2 & 99.8$\pm$0.2 & 84.7$\pm$0.3 & 68.6$\pm$0.3 & 70.0$\pm$0.4 & 84.3\\
        MADA \cite{pei2018multi}& 90.0$\pm$0.1 & 97.4$\pm$0.1 & 99.6$\pm$0.1 & 87.8$\pm$0.2 & 70.3$\pm$0.3 & 66.4$\pm$0.3 & 85.2\\
        MCD \cite{saito2018maximum}& 89.6$\pm$0.2 & 98.5$\pm$0.1 & \textbf{100.0}$\pm$.0 & \textbf{91.3}$\pm$0.2 & 69.6$\pm$0.1 & 70.8$\pm$0.3 & 86.6\\
        CDAN \cite{long2018conditional}& \textbf{93.1}$\pm$0.2 & 98.2$\pm$0.2 & \textbf{100.0}$\pm$.0 & 89.8$\pm$0.3 & 70.1$\pm$0.4 & 68.0$\pm$0.4 & 86.6\\
	  \midrule 
        \textbf{MLA-DA} & 92.8$\pm$0.2 & \textbf{98.9}$\pm$0.2 & \textbf{100.0}$\pm$.0 & 91.2$\pm$0.4 & \textbf{74.7}$\pm$0.4 & \textbf{71.0}$\pm$0.1 & \textbf{88.1}\\
	  \bottomrule 
    \end{tabular}
    \label{Office-31} 
\end{table*}

\subsection{Implementation Details}
We follow the commonly used experiment protocol for unsupervised domain adaptation from \cite{ganin2016domain,long2018conditional}. We report the average accuracies of five independent experiments.
\label{DAvsMA}

We implement our algorithm in \textbf{Pytorch}. For the deep learning experiments, \textbf{ResNet-50} \cite{he2016deep} is adopted as the feature extractor with parameters fine-tuned from the pre-trained ImageNet \cite{russakovsky2015imagenet}. The classifier and metric generator are both 2-layer neural networks with width 1000. For optimization, we use the mini-batch SGD with the momentum 0.9. The minimax problem is implemented by introducing a gradient reversal layer \cite{ganin2016domain}. The learning rate of the classifier, discriminator and metric generator are set 10 times to that of the feature extractor, the value of which is adjusted according to \cite{ganin2016domain}. The batch size is set to 32 in all experiments except Office-Home, which is set to 64.

For hyper-parameters, we fix $\lambda$ as 0.1, $\alpha_0$ as 5 in all the experiments on every transfer task and take the value of $\mu$ to be the number of class $K$ in each dataset. The only hyper-parameter that needs to be adjusted is $\gamma$, which is the weight of our triplet loss. We compare the performance of D$\rightarrow$A and W$\rightarrow$A task on different $\gamma$ as shown in Table \ref{gamma}. We observe see that $\gamma=0.08$ achieves the best performance on two small-to-large transfer tasks by searching in steps of 0.01. And we fixed $\gamma=0.08$ for all the experiments. Additionally, we demonstrate that the dynamic margin is better than constant margin as shown in Table \ref{dynamic}.

\begin{table*}[h]
    \centering 
    \caption{Accuracy (\%) on Office-Home for unsupervised domain adaptation (ResNet-50)} 
    \resizebox{\textwidth}{15mm}{
    \begin{tabular}{cccccccccccccc} 
        \toprule 
        Method & Ar$\rightarrow$Cl & Ar$\rightarrow$Pr & Ar$\rightarrow$Rw & Cl$\rightarrow$Ar & Cl$\rightarrow$Pr & Cl$\rightarrow$Rw & Pr$\rightarrow$Ar & Pr$\rightarrow$Cl &  Pr$\rightarrow$Rw &  Rw$\rightarrow$Ar &  Rw$\rightarrow$Cl &  Rw$\rightarrow$Pr &  Avg \\ 
        \midrule 
        Resnet-50 \cite{he2016deep}& 42.5 & 50.0 & 58.0 & 37.4 & 41.9 & 46.2 & 38.5 & 42.4 & 60.4 & 53.9 & 41.2 & 59.9 & 47.7\\
        DAN \cite{long2015learning}& 43.6 & 57.0 & 67.9 & 45.8 & 56.5 & 60.4 & 44.0 & 43.6 & 67.7 & 63.1 & 51.5 & 74.3 & 56.3\\ 
        DANN \cite{ganin2016domain}& 45.6 & 59.3 & 70.1 & 47.0 & 58.5 & 60.9 & 46.1 & 43.7 & 68.5 & 63.2 & 51.8 & 76.8 & 57.6\\ 
        JAN \cite{long2017deep}& 45.9 & 61.2 & 68.9 & 50.4 & 59.7 & 61.0 & 45.8 & 43.4 & 70.3 & 63.9 & 52.4 & 76.8 & 58.3\\ 
        CDAN \cite{long2018conditional}& 49.0 & 69.3 & 74.5 & 54.4 & 66.0 & 68.4 & 55.6 & 48.3 & 75.9 & 68.4 & 55.4 & 80.5 & 63.8\\ 
	  \midrule 
        \textbf{MLA-DA}& \textbf{54.9} & \textbf{70.4} & \textbf{75.8} & \textbf{58.9} & \textbf{68.0} & \textbf{69.3} & \textbf{59.1} & \textbf{53.1} & \textbf{78.9} & \textbf{70.1} & \textbf{60.5} & \textbf{82.0} & \textbf{66.8}\\ 
	  \bottomrule 
    \end{tabular}}
    \label{Office-Home} 
\end{table*}

\begin{table*}[h]
    \centering 
    \caption{Accuracy (\%) on ImageCLEF-DA for unsupervised domain adaptation (ResNet-50)} 
    \begin{tabular}{cccccccc} 
        \toprule 
        Method & I$\rightarrow$P & P$\rightarrow$I & I$\rightarrow$C & C$\rightarrow$I & C$\rightarrow$P & P$\rightarrow$C & Avg \\ 
        \midrule 
        Resnet-50 \cite{he2016deep}& 74.8$\pm$0.3 & 83.9$\pm$0.1 & 91.5$\pm$0.3 & 78.0$\pm$0.2 & 65.5$\pm$0.3 & 91.2$\pm$0.3 & 80.7\\ 
        DAN \cite{long2015learning}& 74.5$\pm$0.4 & 82.2$\pm$0.2 & 92.8$\pm$0.2 & 86.3$\pm$0.4 & 69.2$\pm$0.4 & 89.8$\pm$0.4 & 82.5\\ 
        DANN \cite{ganin2016domain}& 75.0$\pm$0.6 & 86.0$\pm$0.3 & 96.2$\pm$0.4 & 87.0$\pm$0.5 & 74.3$\pm$0.5 & 91.5$\pm$0.6 & 85.0\\ 
        JAN \cite{long2017deep}& 76.8$\pm$0.4 & 88.0$\pm$0.2 & 94.7$\pm$0.2 & 89.5$\pm$0.3 & 74.2$\pm$0.3 & 91.7$\pm$0.3 & 85.8\\
        MADA \cite{pei2018multi}& 75.0$\pm$0.3 & 87.9$\pm$0.2 & 96.0$\pm$0.3 & 88.8$\pm$0.3 & 75.2$\pm$0.2 & 92.2$\pm$0.3 & 85.8\\
        CDAN \cite{long2018conditional}& 76.7$\pm$0.3 & 90.6$\pm$0.3 & \textbf{97.0}$\pm$0.4 & 90.5$\pm$0.4 & 74.5$\pm$0.3 & 93.5$\pm$0.4 & 87.1\\
	  \midrule 
        \textbf{MLA-DA} & \textbf{79.0}$\pm$0.2 & \textbf{91.3}$\pm$0.2 & 96.5$\pm$0.2 & \textbf{91.5}$\pm$0.2 & \textbf{77.2}$\pm$0.2 & \textbf{94.5}$\pm$0.2 & \textbf{88.3}\\
	  \bottomrule 
    \end{tabular} 
    \label{ImageCLEF-DA} 
\end{table*}

\begin{table}[h]
    \centering 
    \caption{Accuracy (\%) on VisDA-2017 for unsupervised domain adaptation (ResNet-101)} 
    \begin{tabular}{cc} 
        \toprule 
        Method & Synthetic$\rightarrow$Real \\ 
        \midrule 
        Resnet-50 \cite{he2016deep}&  52.4\\
        RevGrad \cite{ganin2015unsupervised}& 57.4\\ 
        DAN \cite{long2015learning}& 61.1\\ 
        MCD \cite{saito2018maximum}& 71.9\\ 
        CDAN \cite{long2018conditional}& 73.7\\ 
	  \midrule 
        \textbf{MLA-DA}& \textbf{75.5}\\ 
	  \bottomrule 
    \end{tabular}
    \label{VisDA-2017} 
\end{table}

\newcommand{\tabincell}[2]{\begin{tabular}{@{}#1@{}}#2\end{tabular}}
\begin{table*}
    \centering 
    \caption{Comparison of cosine distance between critical sample pairs on four datasets} 
	\scalebox{0.87}{
    \begin{tabular}{|c|c|c|c|c|}
        \hline
        Dataset & Method & Anchor & The farthest positive sample & The nearest negative sample\\ 
        \hline 
		\multirow{16}{*}{\tabincell{c}{Office-31\\(D$\rightarrow$A)}}
        &\multirow{8}{*}{MLA-DA} 
		& label: notebook & label: notebook & label: ring binder\\
		&& PR of a ring binder: 0.424 & Cosine distance: 1.312 &Cosine distance: 0.731\\
      	& &\includegraphics[width=0.12\linewidth]{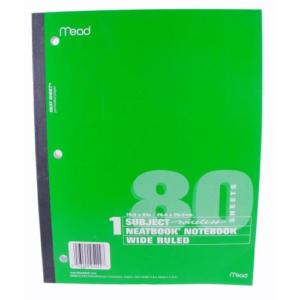} &
	    \includegraphics[width=0.12\linewidth]{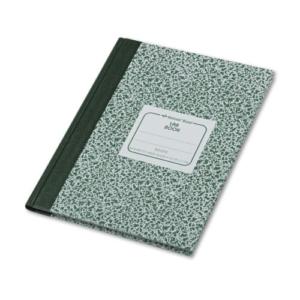}&
	    \includegraphics[width=0.12\linewidth]{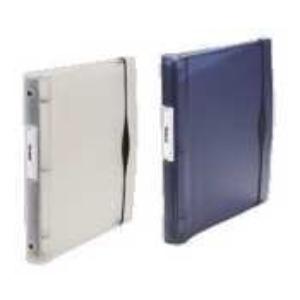}\\
        \cline{2-5}
		&\multirow{8}{*}{DA} 
		& label: punchers & label: punchers &label: trash can\\
		&& PR of a trash can: 0.420 & Cosine distance: 1.634 &  Cosine distance: 0.341\\
      	& &\includegraphics[width=0.14\linewidth]{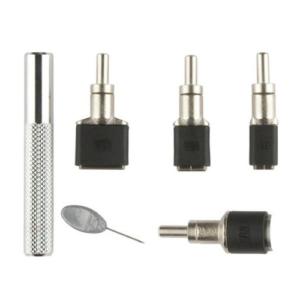} &
	    \includegraphics[width=0.12\linewidth]{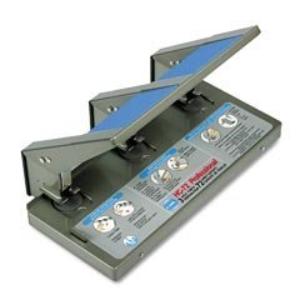}&
	    \includegraphics[width=0.12\linewidth]{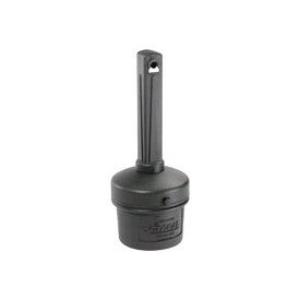}\\
        \hline 

		\multirow{16}{*}{\tabincell{c}{Office-Home\\(Ar$\rightarrow$Cl)}}
        &\multirow{8}{*}{MLA-DA} 
		& label: Knives & label: Knives & label: screwdriver\\
		&& PR of a ring screwdriver: 0.400 & Cosine distance: 1.358 & Cosine distance: 0.496\\
      	& &\includegraphics[width=0.15\linewidth]{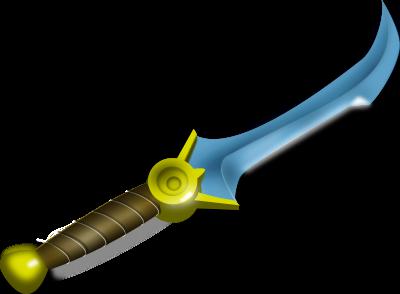} &
	    \includegraphics[width=0.13\linewidth]{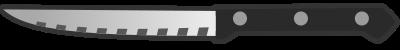}&
	    \includegraphics[width=0.12\linewidth]{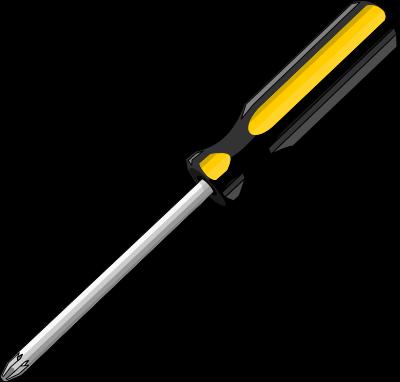}\\
        \cline{2-5}
		&\multirow{8}{*}{DA} 
		& label: computer & label: computer & label: speaker\\
		&& PR of a speaker: 0.437 & Cosine distance: 1.528 & Cosine distance: 0.599\\
      	& &\includegraphics[width=0.12\linewidth]{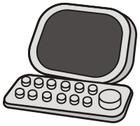} &
	    \includegraphics[width=0.12\linewidth]{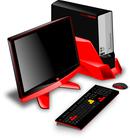}&
	    \includegraphics[width=0.09\linewidth]{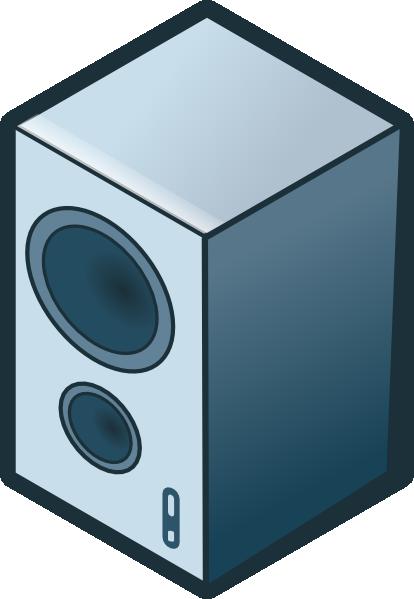}\\
        \hline 

		\multirow{16}{*}{\tabincell{c}{ImageCLEF-DA\\(I$\rightarrow$P)}}
        &\multirow{8}{*}{MLA-DA} 
		& label: bicycle & label: bicycle & label: bus\\
		&& PR of a bus: 0.294 & Cosine distance: 1.468 & Cosine distance: 0.497\\
      	& &\includegraphics[width=0.14\linewidth]{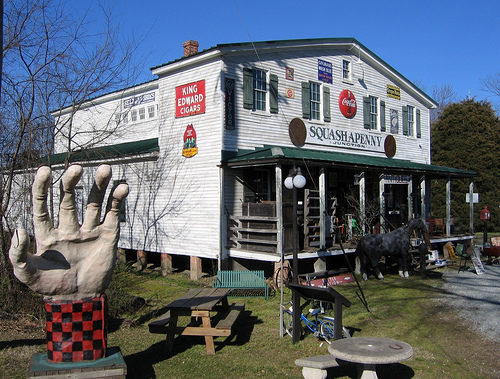} &
	    \includegraphics[width=0.14\linewidth]{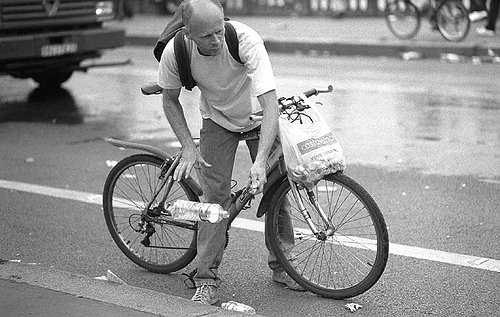}&
	    \includegraphics[width=0.14\linewidth]{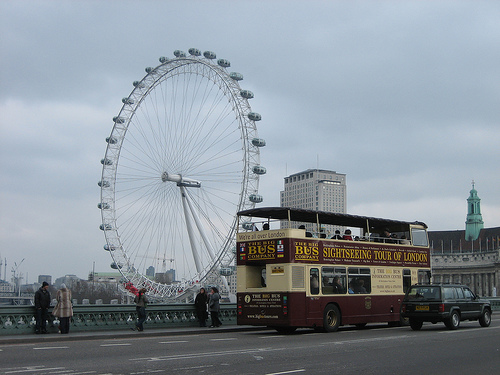}\\
        \cline{2-5}
		&\multirow{8}{*}{DA} 
		& label: bus & label: bus & label: aeroplane\\
		&& PR of a aeroplane: 0.316 & Cosine distance: 1.452 & Cosine distance: 0.205\\
      	& &\includegraphics[width=0.16\linewidth]{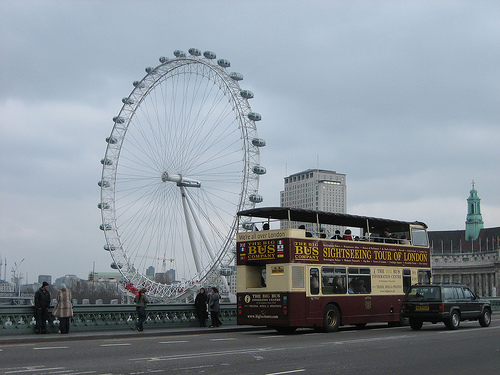} &
	    \includegraphics[width=0.15\linewidth]{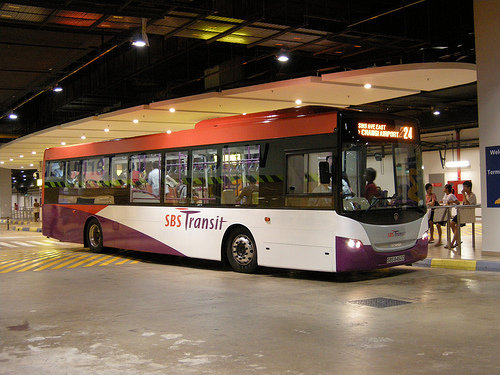}&
	    \includegraphics[width=0.16\linewidth]{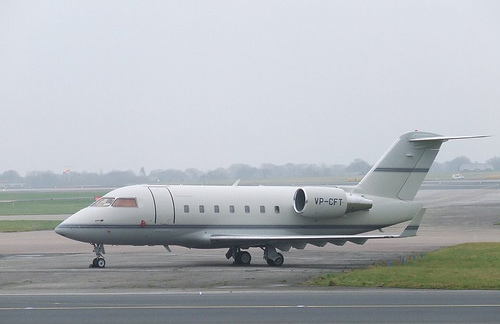}\\
        \hline 

		\multirow{16}{*}{\tabincell{c}{VisDA-2017\\(Syn$\rightarrow$Real)}}
        &\multirow{8}{*}{MLA-DA} 
		& label: bus & label: bus & label: train\\
		&& PR of a train: 0.376 & Cosine distance: 0.920 & Cosine distance: 0.716\\
      	& &\includegraphics[width=0.16\linewidth]{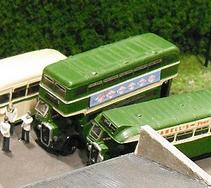} &
	    \includegraphics[width=0.13\linewidth]{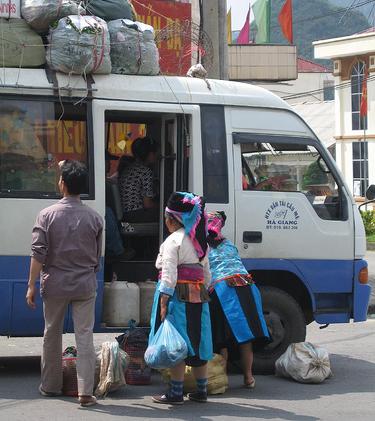}&
	    \includegraphics[width=0.2\linewidth]{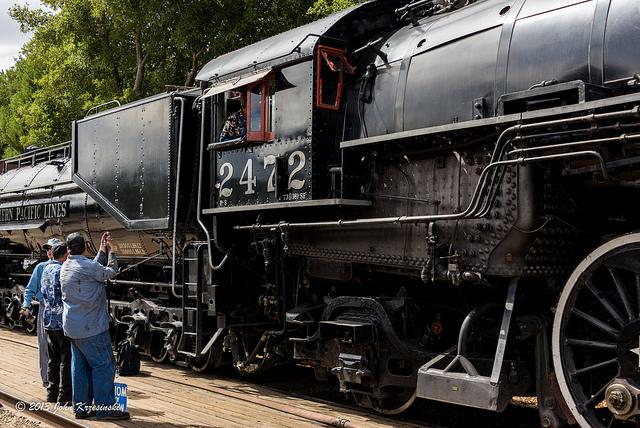}\\
        \cline{2-5}
		&\multirow{10}{*}{DA} 
		& label: car & label: car & label: truck\\
		&& PR of a truck: 0.451 & Cosine distance: 1.334 & Cosine distance: 0.207\\
      	& &\includegraphics[width=0.24\linewidth]{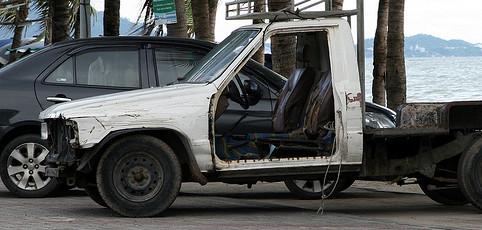} &
	    \includegraphics[width=0.10\linewidth]{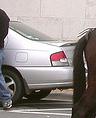}&
	    \includegraphics[width=0.22\linewidth]{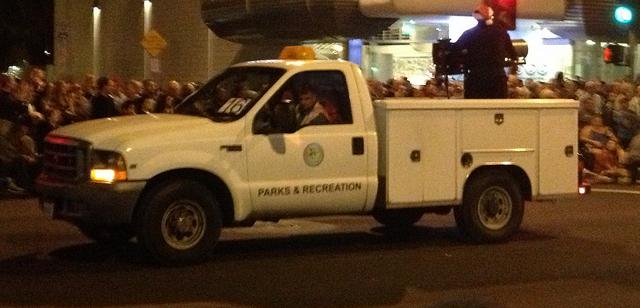}\\
        \hline
    \end{tabular}}
    \label{distance} 
\end{table*}

\subsection{Metric-Learning-Assisted Domain Adaptation vs. Domain Alignment}
In this subsection, we demonstrate the impact of MLA-DA from three perspectives: feature distribution, target accuracy (classification accuracy on target dataset) and triplet loss. As shown in Fig. \ref{t-SNE}, experiments on the same transfer task (VisDA) are compared with different loss functions. For a fair comparison, the target entropy loss $\mathcal{L}_{E}$ is not considered in these three experiments. We can see that the embedded features can be well separated by the use of our triplet loss. 

Then, we compare the target accuracy and triplet loss on the two small-to-large transfer tasks (D$\rightarrow$A, W$\rightarrow$A) in Fig. \ref{acc_loss}. We can see that the target accuracy of both challenging tasks can be improved by minimizing the proposed triplet loss $\mathcal{L}_T$, while it is difficult to improve the performance by using only domain alignment loss $\mathcal{L}_D$. \textbf{Interestingly}, minimizing target entropy loss can reduce the triplet loss and improve the target accuracy. The decision boundary is forced to be far away from the region with dense samples by minimizing target entropy loss \cite{grandvalet2005semi}, and the interval between different classes increases correspondingly. This phenomenon shows that our triplet loss is effective and universal for domain adaptation, but not explored in the previous works. 

Meanwhile, we find that aligned target features are indeed not discriminative enough for classification as shown in Fig. \ref{D-A_acc} and \ref{W-A_acc}. On these two small-to-large transfer tasks, source samples are insufficient for generalization and the deep model tends to be overfitting. Our experiments demonstrate that low source risk and source-target feature alignment does not imply low target risk when the source domain is a small dataset.

From Fig. \ref{acc_loss}, we have the following conclusions on two small-to-large transfer tasks (D$\rightarrow$A, W$\rightarrow$A):
\begin{itemize}
    \item The use of $\mathcal{L}_C+\mathcal{L}_D$ does not improve target accuracy compared with $\mathcal{L}_C$, domain alignment does not work.
    \item The use of $\mathcal{L}_C+\gamma \mathcal{L}_T$ is better than $\mathcal{L}_C+\mathcal{L}_D$, the use of triplet loss is more effective than domain alignment loss.
    \item The use of $\mathcal{L}_C+\mathcal{L}_D+\gamma \mathcal{L}_T$ is much better than $\mathcal{L}_C+\gamma \mathcal{L}_T$ and $\mathcal{L}_C+\mathcal{L}_D$, the triplet loss works well together with domain alignment loss.
    \item The use of $\mathcal{L}_C+\mathcal{L}_D+\gamma \mathcal{L}_T+\lambda \mathcal{L}_E$ can achieve the state-of-the-art performance.
\end{itemize}

\subsection{Comparison of critical sample pairs}
In this subsection, we compare the cosine distance between critical sample pairs of MLA-DA and DA on all the four datasets. The critical sample pairs consist of two parts: the farthest positive sample and the nearest negative sample. For each dataset, we focus on the tasks: D$\rightarrow$A in Office-31, Ar$\rightarrow$Cl in Office-Home, I$\rightarrow$P in ImageCLEF-DA and Synthetic$\rightarrow$Real in VisDA-2017. Firstly, we make inference over the trained model to obtain the embeddings and classification results of the target task samples. Then, we locate the most uncertain sample, which most likely to be misclassified. In the third column, we show the most uncertain sample, its second largest probability of prediction and its true label. In the fourth and fifth column, we find the farthest samples of the same category and the nearest samples of different categories by calculating the cosine distance of the original feature space. In the fifth column, we also show the true labels of the negative samples. The cosine distance is defined as:
{\color{blue}{
\begin{align}
d_c(\bm{f},\bm{f'})&=1-\cos (\theta)
\nonumber\\ &=1-\frac{\bm{f} \cdot \bm{f'}}{\|\bm{f}\|\|\bm{f'}\|} 
=1-\frac{\sum_{i=1}^{n} \bm{f}_{i} \times \bm{f'}_{i}}{\sqrt{\sum_{i=1}^{n}\left(\bm{f}_{i}\right)^{2}} \times \sqrt{\sum_{i=1}^{n}\left(\bm{f'}_{i}\right)^{2}}},
\label{cosine distance}
\end{align}where $\bm{f}=f(\bm{x}), \bm{f'}=f(\bm{x'})$ and $\cos (\theta)$ is the cosine similarity between $\bm{f}$ and $\bm{f'}$.}} We show the cosine distance between the positive pairs and the negative pairs in the last two columns.

The conclusion of the feature separation result of Table \ref{distance} can be summarized as follows:
\begin{itemize}
    \item These uncertain samples have large probabilities of predicting as wrong categories. This conclusion is consistent with the discovery of Fig. \ref{fsubpeak}.
    \item Samples in the third and fifth columns are visually similar, even if they belong to two different categories. These similar negative samples confuse the classifier. Similarly, the completely different positive samples also make the classifier not confident on its prediction.
    \item Generally, in MLA-DA, the minimum distance of negative pairs in feature space is greater than DA, and the maximum distance of the positive pairs is smaller than in the DA method. Correspondingly, in MLA-DA, the probability that an uncertain sample is misclassified into other categories is smaller than in the DA method. This shows that MLA-DA successfully separate target features from the decision boundaries indirectly by the use of proposed triplet loss.
\end{itemize}

\begin{table*}[h]
    \centering 
    \caption{{\color{blue}{Ablation experiments on Office-31 for unsupervised domain adaptation (ResNet-50)}}} 
    \begin{tabular}{cccccccc} 
        \toprule 
        Loss function combinations & A$\rightarrow$W & D$\rightarrow$W & W$\rightarrow$D & A$\rightarrow$D & D$\rightarrow$A & W$\rightarrow$A & Avg \\ 
        \midrule 
        $\mathcal{L}_C$ & 68.4$\pm$0.2 & 96.7$\pm$0.1 & 99.3$\pm$0.1 & 68.9$\pm$0.2 & 62.5$\pm$0.3 & 62.7$\pm$0.3 & 76.2\\ 
	  \midrule 
        $\mathcal{L}_C+\mathcal{L}_D$ & 83.0$\pm$0.1 & 97.8$\pm$0.1 & 99.8$\pm$0.2 & 80.6$\pm$0.2 & 62.7$\pm$0.3 & 60.0$\pm$0.1 & 80.7\\
        $\mathcal{L}_C+\bm{\gamma \mathcal{L}_T}$ & 81.4$\pm$0.3 & 98.3$\pm$0.2 & \textbf{100.0}$\pm$.0 & 85.0$\pm$0.1 & 64.0$\pm$0.3 & 62.9$\pm$0.2 & 81.9\\
        $\mathcal{L}_C+\mathcal{L}_D+\bm{\gamma \mathcal{L}_T}$ & 82.8$\pm$0.2 & 97.9$\pm$0.1 & \textbf{100.0}$\pm$.0 & 81.7$\pm$0.3 & 62.6$\pm$0.3 & 60.7$\pm$0.2 & 81.0\\
	  \midrule 
        $\mathcal{L}_C+\mathcal{L}_D+\lambda \mathcal{L}_E$ & 91.6$\pm$0.1 & 98.7$\pm$0.1 & 99.9$\pm$0.1 & 89.5$\pm$0.3 & 73.5$\pm$0.6 & 67.0$\pm$0.1 & 86.7\\
        $\mathcal{L}_C+\mathcal{L}_D+\lambda \mathcal{L}_E+\bm{\gamma \mathcal{L}_T}$ & \textbf{92.8} $\pm$0.2 & \textbf{98.9}$\pm$0.2 & \textbf{100.0}$\pm$.0 & \textbf{91.2}$\pm$0.4 & \textbf{74.7}$\pm$0.4 & \textbf{71.0}$\pm$0.1 & \textbf{88.1}\\
	  \bottomrule 
    \end{tabular}
    \label{Ablation Study} 
\end{table*}

\begin{table*}[h]
    \centering 
    \caption{Accuracy (\%) on downsampled source domains (ResNet-50)} 
    \begin{tabular}{cccccccccc} 
        \toprule 
        Method & D/2$\rightarrow$A & D/4$\rightarrow$A & W/2$\rightarrow$A & W/4$\rightarrow$A & Ar/2$\rightarrow$Cl & Ar/4$\rightarrow$Cl & Pr/2$\rightarrow$Cl & Pr/4$\rightarrow$Cl & Avg\\
	   \midrule 
	   Mean Labeled/Class& 8 & 4 & 13 & 7 & 19 & 10 & 34 & 17 & 14 \\
	   Source Size& 255 & 134 & 405 & 208 & 1228 & 629 & 2238 & 1132 & 779 \\
        Target Size& 2817 & 2817 & 2817 & 2817 & 4365 & 4365 & 4365 & 4365 & 3591 \\
        \midrule 
        Resnet-50 \cite{he2016deep}& 60.7 & 58.3 & 62.3 & 61.2 & 42.4 & 38.6 & 41.8 & 41.0 & 50.8 \\
        DA\cite{ganin2016domain}& 62.4 & 58.7 & 62.3 & 60.1 & 43.2 & 39.8 & 43.5 & 41.4 & 51.4 \\ 
        CDAN \cite{long2018conditional}& 68.8 & 62.9 & 66.9 & 64.5 & 43.9 & 36.1 & 46.1 & 43.4 & 54.1 \\
	  \midrule 
        \textbf{MLO-DA}& 65.3 & 61.4 & 65.4 & \textbf{65.9} & 46.0 & \textbf{41.9} & 45.5 & 45.7 & 54.6 \\ 
	  \textbf{MLA-DA}& \textbf{69.9} & \textbf{63.4} & \textbf{68.1} & \textbf{65.9} & \textbf{46.1} & 41.6 & \textbf{49.6} & \textbf{46.4} & \textbf{56.3} \\
	  \bottomrule 
    \end{tabular}
    \label{WSDA Challenge} 
\end{table*}

\subsection{Results on Benchmarks}
The result on Office-31 are reported in Table \ref{Office-31}. We could see that MLA-DA achieves state-of-the-art accuracies on four of six transfer tasks. We note that in previous works, CDAN performs a little better for large-to-small transfer tasks (A$\rightarrow$W, A$\rightarrow$D). Nevertheless, our algorithm outperforms on two small-to-large transfer tasks (D$\rightarrow$A, W$\rightarrow$A) and achieves higher performance than well-known methods in the previous works, demonstrating the effectiveness and universality of MLA-DA.

Table \ref{Office-Home}, \ref{VisDA-2017} and \ref{ImageCLEF-DA} present the accuracies of our algorithm on Office-Home, VisDA2017 and ImageCLEF-DA datasets. MLA-DA achieves the best performance on almost all transfer tasks. This validates the effectiveness and universality of MLA-DA.

{\color{blue}{
\subsection{Ablation Study}
In this section, we do ablation study on MLA-DA to show the effect of different loss function combinations. As shown in Table \ref{Ablation Study}, The effect of various combinations of loss functions are investigated on Office-31 dataset. We have the following comments as shown in Table \ref{Ablation Study}:

$\bm{\mathcal{L}_C+\mathcal{L}_D\ vs.\ \mathcal{L}_C+\mathcal{L}_T}$. The use of triplet loss $\mathcal{L}_T$ can improve the performance without domain alignment loss $\mathcal{L}_D$, and achieve better performance than the use of $\mathcal{L}_D$. 

$\bm{\mathcal{L}_C+\mathcal{L}_T\ vs.\ \mathcal{L}_C+\mathcal{L}_D+\gamma \mathcal{L}_T}$. Without target entropy loss $\mathcal{L}_E$, the use of domain alignment loss $\mathcal{L}_D$ leads to a decline in performance. While the use of triplet loss $\mathcal{L}_T$ improves the performance in the $\bm{\mathcal{L}_C+\mathcal{L}_D\ vs.\ \mathcal{L}_C+\mathcal{L}_D+\gamma \mathcal{L}_T}$ group.

$\bm{\mathcal{L}_C+\mathcal{L}_D+\lambda \mathcal{L}_E\ vs.\ \mathcal{L}_C+\mathcal{L}_D+\lambda \mathcal{L}_E+\gamma \mathcal{L}_T}$. With the use of target entropy loss $\mathcal{L}_E$ and domain alignment loss $\mathcal{L}_D$, the use of our triplet loss brings considerable performance improvements. This performance is comparable to the state-of-the-art works \cite{DBLP:conf/cvpr/ZhangTJT19,DBLP:conf/cvpr/ChangYSKH19}, and the performance gap is no more than 0.3\%.
}}

\subsection{Robustness Analysis}
\label{WSDA}
\begin{algorithm}[htb]
\caption{Noise Generating via VAT}
\label{VAT}
{\bf Input:}\\
$\mathcal{X}_T=\{ \textbf{x}_{t}^{\left( i \right)}\}_{i=1}^{n_t}$: target training sample set;\\
$F: \mathcal{X} \rightarrow \mathbb{R}^n$: embedding function parameterized by $\theta _F$;\\
$C: \mathbb{R}^n \rightarrow \mathbb{R}^k$: embedding classifier parameterized by $\theta _C$;\\
$I_n$: the intensity of noise.
\begin{algorithmic}[1]
\For{each batch $\mathcal{X}^{batch}_{T}$ in $\mathcal{X}_T$}
	\State generate normally distributed random matrix $\bm{N}$ with the same size as $\mathcal{X}^{batch}_{T}$;
	\State estimate the probability that the sample belongs to each category $\hat{P}(Y|X)=C(F(\mathcal{X}^{batch}_{T}))$ and $\hat{P}(Y|X+\bm{N})=C(F(\mathcal{X}^{batch}_{T}+\bm{N}))$;
	\State calculate the pseudo label $\mathcal{Y}^{pseudo}_{T}=\mathop{\arg\max}_{y} \hat{P}(Y|X)$ of target data;
	\State calculate cross-entropy loss $\mathcal{L}^{pseudo}_C=L(C(F(\mathcal{X}^{batch}_{T}+\bm{N})),\mathcal{Y}^{pseudo}_{T})$;
	\State calculate the gradient $\nabla_{\bm{N}} \mathcal{L}^{pseudo}_C$;
	\State generate noisy data $\mathcal{X}^{noisy}_{T}\leftarrow \mathcal{X}^{batch}_{T}+I_n \nabla_{\bm{N}} \mathcal{L}^{pseudo}_C$.
\EndFor
\end{algorithmic}
{\bf Output:}
$\mathcal{X}^{noisy}_{T}$.
\end{algorithm}
By taking a subset of source dataset, we can create a small source dataset of reduced size. We choose the four most challenging transfer tasks in classic benchmarks: D$\rightarrow$A, W$\rightarrow$A in Office-31 and Ar$\rightarrow$Cl, Pr$\rightarrow$Cl in Office-Home. To fairly compare the impact of each loss, various possible combination methods are compared in Table \ref{WSDA Challenge}, including Domain Alignment (\textbf{DA}: $\mathcal{L}_C+\mathcal{L}_D$), Metric-Learning-Only Domain Adaptation (\textbf{MLO-DA}: $\mathcal{L}_C+\mathcal{L}_D+\gamma \mathcal{L}_T$) and Metric-Learning-Assisted Domain Adaptation (\textbf{MLA-DA}: $\mathcal{L}_C+\mathcal{L}_D+\gamma \mathcal{L}_T+ \lambda \mathcal{L}_{E}$). The experimental result shows that DA does not work well with the decrease of source samples. While MLA-DA achieves performance improvement in all the robustness testing tasks. We find that MLA-DA is robust during the reduction of source size, and it has more advantages in the small source domain scenario.

\begin{figure}[h]
	\centering
	\subfigure[Target data ($I_n=0$)]{
		\begin{minipage}[c]{0.40\linewidth}
			\centering
			\includegraphics[width=0.8\linewidth]{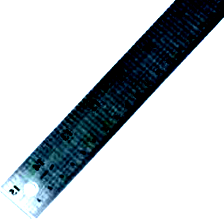}
			\label{target data}
		\end{minipage}}
	\subfigure[Gradient of noise ($\nabla_{\bm{N}} \mathcal{L}^{pseudo}_C$)]{
		\begin{minipage}[c]{0.40\linewidth}
			\centering
			\includegraphics[width=0.8\linewidth]{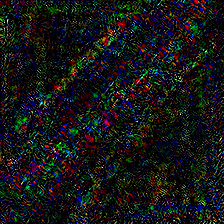}
			\label{gradient of noise}
		\end{minipage}}
	\subfigure[Noisy data ($I_n=3.5$)]{
		\begin{minipage}[c]{0.40\linewidth}
			\centering
			\includegraphics[width=0.8\linewidth]{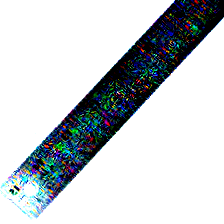}
			\label{noisy data 3.5}
		\end{minipage}}
	\subfigure[Noisy data ($I_n=5$)]{
		\begin{minipage}[c]{0.40\linewidth}
			\centering
			\includegraphics[width=0.8\linewidth]{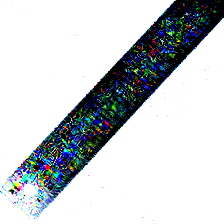}
			\label{noisy data 5}
		\end{minipage}}
	\caption{An example of the target data \ref{target data}, the gradient of noise \ref{gradient of noise} and the noisy data \ref{noisy data 3.5} and \ref{noisy data 5}. The gradient of noise and the noisy data are generated by Algorithm \ref{VAT}.}
	\label{noise}
\end{figure}

In addition to reducing the size of source domain, we also test the anti-noise ability of the classifier. To test accurately, we use virtual adversarial training (VAT) \cite{DBLP:journals/pami/MiyatoMKI19} to generate noise. The algorithm of noise generating via VAT is showm as Algorithm \ref{VAT}. One of the target data $\mathcal{X}_T$, the gradient $\nabla_{\bm{N}} \mathcal{L}^{pseudo}_C$ and noisy data $\mathcal{X}^{noisy}_{T}$ are shown in Fig. \ref{noise}. The noisy data generated by Algorithm \ref{VAT} are used to test the robustness of the classifier. The five methods are compared in Table \ref{robustness}. The experimental results show that MLA-DA has the best anti-noise ability compared with other methods.

\begin{table}[h]
    \centering 
    \caption{Accuracy (\%) on noisy data (ResNet-50)} 
	\scalebox{0.88}{
    \begin{tabular}{|c|c|c|c|c|c|c|}
        \hline
        Method & $I_n$ & D$\rightarrow$A & W$\rightarrow$A & Ar$\rightarrow$Cl & Pr$\rightarrow$Cl & Avg\\ 
        \hline 
		\multirow{3.25}{*}{\tabincell{c}{Resnet-50 \cite{he2016deep}}}
        	&\multirow{1.25}{*}{0} 
		& 62.5 & 62.7 & 42.5 & 42.4 & 52.5\\
        	\cline{2-7}
		&\multirow{1.25}{*}{3.5}
		& 38.9 & 41.3 & 31.1 & 29.2 & 35.1\\
		\cline{2-7}
		&\multirow{1.25}{*}{5}
		& 37.0 & 38.7 & 31.8 & 28.2 & 33.9\\
        \hline
		\multirow{3.25}{*}{\tabincell{c}{DA\cite{ganin2016domain}}}
        	&\multirow{1.25}{*}{0} 
		& 64.2 & 62.1 & 45.6 & 43.7 & 56.2\\
        	\cline{2-7}
		&\multirow{1.25}{*}{3.5}
		& 40.5 & 42.9 & 32.0 & 33.0 & 37.1\\
		\cline{2-7}
		&\multirow{1.25}{*}{5}
		& 39.3 & 39.4 & 30.3 & 31.6 & 35.2\\
        \hline
		\multirow{3.25}{*}{\tabincell{c}{CDAN \cite{long2018conditional}}}
        	&\multirow{1.25}{*}{0} 
		& 70.1 & 68.0 & 49.0 & 48.3 & 58.9\\
        	\cline{2-7}
		&\multirow{1.25}{*}{3.5}
		& 33.7 & 34.4 & 23.0 & 21.8 & 28.2\\
		\cline{2-7}
		&\multirow{1.25}{*}{5}
		& 33.8 & 22.8 & 22.3 & 17.8 & 24.2\\
        \hline
		\multirow{3.25}{*}{\tabincell{c}{MLO-DA}}
        	&\multirow{1.25}{*}{0} 
		& 68.2 & 66.1 & 49.2 & 48.0 & 57.9\\
        	\cline{2-7}
		&\multirow{1.25}{*}{3.5}
		& 50.5 & 48.2 & 36.1 & 35.8 & 42.7\\
		\cline{2-7}
		&\multirow{1.25}{*}{5}
		& 46.5 & 47.7 & 35.7 & 34.3 & 41.1\\
        \hline
		\multirow{3.25}{*}{\tabincell{c}{MLA-DA}}
        	&\multirow{1.25}{*}{0} 
		& 74.7 & 71.0 & 54.9 & 53.1 & 63.4\\
        	\cline{2-7}
		&\multirow{1.25}{*}{3.5}
		& 61.4 & 59.3 & 47.4 & 46.3 & 53.6\\
		\cline{2-7}
		&\multirow{1.25}{*}{5}
		& 57.3 & 56.9 & 44.2 & 44.5 & 50.7\\
        \hline
    \end{tabular}}
    \label{robustness} 
\end{table}

\section{Conclusion}
{\color{blue}{In this paper, we explore the relationship between the second largest probability of a target sample's prediction and its distance to the decision boundary. Based on this relationship, we propose a novel mechanism to adaptively adjust the margin in the triplet loss according to target predictions.}} We {\color{blue}{further}} propose a Metric-Learning-Assisted Domain Adaptation (MLA-DA) by using {\color{blue}{the}} triplet loss, which can address the limitation of domain alignment and obtain a more robust classifier for unsupervised domain adaptation. We show that the use of dynamic margin in triplet loss is beneficial. Extensive experimental results demonstrate the effectiveness and universality of MLA-DA.

\section*{Acknowledgments}
This work was supported in part by the Natural Science Foundation of China under Grant 61671252, 61571233 and 61901229; the Natural Science Research of Higher Education Institutions of Jiangsu Province under Grant 19KJB510008.

\bibliographystyle{elsarticle-num}
\bibliography{Manuscript}

\begin{thebibliography}{10}
\expandafter\ifx\csname url\endcsname\relax
  \def\url#1{\texttt{#1}}\fi
\expandafter\ifx\csname urlprefix\endcsname\relax\def\urlprefix{URL }\fi
\expandafter\ifx\csname href\endcsname\relax
  \def\href#1#2{#2} \def\path#1{#1}\fi

\bibitem{caicedo2019nucleus}
J.~C. Caicedo, A.~Goodman, K.~W. Karhohs, B.~A. Cimini, J.~Ackerman,
  M.~Haghighi, C.~Heng, T.~Becker, M.~Doan, C.~McQuin, et~al., Nucleus
  segmentation across imaging experiments: the 2018 data science bowl, Nature
  methods (2019) 1--7.

\bibitem{kanazawa2019learning}
A.~Kanazawa, J.~Y. Zhang, P.~Felsen, J.~Malik, Learning 3d human dynamics from
  video, in: Proceedings of the IEEE Conference on Computer Vision and Pattern
  Recognition, 2019, pp. 5614--5623.

\bibitem{pan2009survey}
S.~J. Pan, Q.~Yang, A survey on transfer learning, IEEE Transactions on
  knowledge and data engineering 22~(10) (2009) 1345--1359.

\bibitem{bashivan2019neural}
P.~Bashivan, K.~Kar, J.~J. DiCarlo, Neural population control via deep image
  synthesis, Science 364~(6439) (2019) eaav9436.

\bibitem{quionero2009dataset}
J.~Quionero-Candela, M.~Sugiyama, A.~Schwaighofer, N.~D. Lawrence, Dataset
  shift in machine learning, The MIT Press, 2009.

\bibitem{shu2018dirt}
R.~Shu, H.~H. Bui, H.~Narui, S.~Ermon, A dirt-t approach to unsupervised domain
  adaptation, arXiv preprint arXiv:1802.08735 (2018).

\bibitem{yosinski2014transferable}
J.~Yosinski, J.~Clune, Y.~Bengio, H.~Lipson, How transferable are features in
  deep neural networks?, in: Advances in neural information processing systems,
  2014, pp. 3320--3328.

\bibitem{long2015learning}
M.~Long, Y.~Cao, J.~Wang, M.~Jordan, Learning transferable features with deep
  adaptation networks, in: International Conference on Machine Learning, 2015,
  pp. 97--105.

\bibitem{ganin2016domain}
Y.~Ganin, E.~Ustinova, H.~Ajakan, P.~Germain, H.~Larochelle, F.~Laviolette,
  M.~Marchand, V.~Lempitsky, Domain-adversarial training of neural networks,
  The Journal of Machine Learning Research 17~(1) (2016) 2096--2030.

\bibitem{tzeng2017adversarial}
E.~Tzeng, J.~Hoffman, K.~Saenko, T.~Darrell, Adversarial discriminative domain
  adaptation, in: Proceedings of the IEEE Conference on Computer Vision and
  Pattern Recognition, 2017, pp. 7167--7176.

\bibitem{saito2018maximum}
K.~Saito, K.~Watanabe, Y.~Ushiku, T.~Harada, Maximum classifier discrepancy for
  unsupervised domain adaptation, in: Proceedings of the IEEE Conference on
  Computer Vision and Pattern Recognition, 2018, pp. 3723--3732.

\bibitem{zhang2019bridging}
Y.~Zhang, T.~Liu, M.~Long, M.~Jordan, Bridging theory and algorithm for domain
  adaptation, in: International Conference on Machine Learning, 2019, pp.
  7404--7413.

\bibitem{ajakan2014domain}
H.~Ajakan, P.~Germain, H.~Larochelle, F.~Laviolette, M.~Marchand,
  Domain-adversarial neural networks, arXiv preprint arXiv:1412.4446 (2014).

\bibitem{kim2017learning}
T.~Kim, M.~Cha, H.~Kim, J.~K. Lee, J.~Kim, Learning to discover cross-domain
  relations with generative adversarial networks, in: Proceedings of the 34th
  International Conference on Machine Learning-Volume 70, JMLR. org, 2017, pp.
  1857--1865.

\bibitem{ma2019gcan}
X.~Ma, T.~Zhang, C.~Xu, Gcan: Graph convolutional adversarial network for
  unsupervised domain adaptation, in: Proceedings of the IEEE Conference on
  Computer Vision and Pattern Recognition, 2019, pp. 8266--8276.

\bibitem{goodfellow2014generative}
I.~Goodfellow, J.~Pouget-Abadie, M.~Mirza, B.~Xu, D.~Warde-Farley, S.~Ozair,
  A.~Courville, Y.~Bengio, Generative adversarial nets, in: Advances in neural
  information processing systems, 2014, pp. 2672--2680.

\bibitem{DBLP:conf/cvpr/ZhangTJT19}
Y.~Zhang, H.~Tang, K.~Jia, M.~Tan, Domain-symmetric networks for adversarial
  domain adaptation, in: {IEEE} Conference on Computer Vision and Pattern
  Recognition, {CVPR}, 2019, pp. 5031--5040.

\bibitem{DBLP:conf/cvpr/ChangYSKH19}
W.~Chang, T.~You, S.~Seo, S.~Kwak, B.~Han, Domain-specific batch normalization
  for unsupervised domain adaptation, in: {IEEE} Conference on Computer Vision
  and Pattern Recognition, {CVPR}, 2019, pp. 7354--7362.

\bibitem{DBLP:conf/cvpr/ChenXHRD0XH19}
C.~Chen, W.~Xie, W.~Huang, Y.~Rong, X.~Ding, Y.~Huang, T.~Xu, J.~Huang,
  Progressive feature alignment for unsupervised domain adaptation, in: {IEEE}
  Conference on Computer Vision and Pattern Recognition, {CVPR}, 2019, pp.
  627--636.

\bibitem{DBLP:conf/iclr/KipfW17}
T.~N. Kipf, M.~Welling, Semi-supervised classification with graph convolutional
  networks, in: 5th International Conference on Learning Representations,
  {ICLR} 2017, 2017.

\bibitem{weinberger2009distance}
K.~Q. Weinberger, L.~K. Saul, Distance metric learning for large margin nearest
  neighbor classification, Journal of Machine Learning Research 10~(Feb) (2009)
  207--244.

\bibitem{xing2003distance}
E.~P. Xing, M.~I. Jordan, S.~J. Russell, A.~Y. Ng, Distance metric learning
  with application to clustering with side-information, in: Advances in neural
  information processing systems, 2003, pp. 521--528.

\bibitem{davis2007information}
J.~V. Davis, B.~Kulis, P.~Jain, S.~Sra, I.~S. Dhillon, Information-theoretic
  metric learning, in: Proceedings of the 24th international conference on
  Machine learning, ACM, 2007, pp. 209--216.

\bibitem{zuo2017distance}
W.~Zuo, F.~Wang, D.~Zhang, L.~Lin, Y.~Huang, D.~Meng, L.~Zhang, Distance metric
  learning via iterated support vector machines, IEEE Transactions on Image
  Processing 26~(10) (2017) 4937--4950.

\bibitem{cheng2017duplex}
G.~Cheng, P.~Zhou, J.~Han, Duplex metric learning for image set classification,
  IEEE Transactions on Image Processing 27~(1) (2017) 281--292.

\bibitem{radenovic2016cnn}
F.~Radenovi{\'c}, G.~Tolias, O.~Chum, Cnn image retrieval learns from bow:
  Unsupervised fine-tuning with hard examples, in: European conference on
  computer vision, Springer, 2016, pp. 3--20.

\bibitem{simo2015discriminative}
E.~Simo-Serra, E.~Trulls, L.~Ferraz, I.~Kokkinos, P.~Fua, F.~Moreno-Noguer,
  Discriminative learning of deep convolutional feature point descriptors, in:
  Proceedings of the IEEE International Conference on Computer Vision, 2015,
  pp. 118--126.

\bibitem{chopra2005learning}
S.~Chopra, R.~Hadsell, Y.~LeCun, et~al., Learning a similarity metric
  discriminatively, with application to face verification, in: CVPR (1), 2005,
  pp. 539--546.

\bibitem{schroff2015facenet}
F.~Schroff, D.~Kalenichenko, J.~Philbin, Facenet: A unified embedding for face
  recognition and clustering, in: Proceedings of the IEEE conference on
  computer vision and pattern recognition, 2015, pp. 815--823.

\bibitem{harwood2017smart}
B.~Harwood, B.~Kumar, G.~Carneiro, I.~Reid, T.~Drummond, et~al., Smart mining
  for deep metric learning, in: Proceedings of the IEEE International
  Conference on Computer Vision, 2017, pp. 2821--2829.

\bibitem{qian2015fine}
Q.~Qian, R.~Jin, S.~Zhu, Y.~Lin, Fine-grained visual categorization via
  multi-stage metric learning, in: Proceedings of the IEEE Conference on
  Computer Vision and Pattern Recognition, 2015, pp. 3716--3724.

\bibitem{grandvalet2005semi}
Y.~Grandvalet, Y.~Bengio, Semi-supervised learning by entropy minimization, in:
  Advances in neural information processing systems, 2005, pp. 529--536.

\bibitem{morerio2018minimal}
P.~Morerio, J.~Cavazza, V.~Murino, Minimal-entropy correlation alignment for
  unsupervised deep domain adaptation (2018).

\bibitem{cariucci2017autodial}
F.~M. Cariucci, L.~Porzi, B.~Caputo, E.~Ricci, S.~R. Bul{\`o}, Autodial:
  Automatic domain alignment layers, in: 2017 IEEE International Conference on
  Computer Vision (ICCV), IEEE, 2017, pp. 5077--5085.

\bibitem{long2018conditional}
M.~Long, Z.~Cao, J.~Wang, M.~I. Jordan, Conditional adversarial domain
  adaptation, in: Advances in Neural Information Processing Systems, 2018, pp.
  1640--1650.

\bibitem{shimodaira2000improving}
H.~Shimodaira, Improving predictive inference under covariate shift by
  weighting the log-likelihood function, Journal of statistical planning and
  inference 90~(2) (2000) 227--244.

\bibitem{do2019theoretically}
T.-T. Do, T.~Tran, I.~Reid, V.~Kumar, T.~Hoang, G.~Carneiro, A theoretically
  sound upper bound on the triplet loss for improving the efficiency of deep
  distance metric learning, in: Proceedings of the IEEE Conference on Computer
  Vision and Pattern Recognition, 2019, pp. 10404--10413.

\bibitem{saenko2010adapting}
K.~Saenko, B.~Kulis, M.~Fritz, T.~Darrell, Adapting visual category models to
  new domains, in: European conference on computer vision, Springer, 2010, pp.
  213--226.

\bibitem{venkateswara2017deep}
H.~Venkateswara, J.~Eusebio, S.~Chakraborty, S.~Panchanathan, Deep hashing
  network for unsupervised domain adaptation, in: Proceedings of the IEEE
  Conference on Computer Vision and Pattern Recognition, 2017, pp. 5018--5027.

\bibitem{DBLP:journals/corr/abs-1710-06924}
X.~Peng, B.~Usman, N.~Kaushik, J.~Hoffman, D.~Wang, K.~Saenko,
  \href{http://arxiv.org/abs/1710.06924}{Visda: The visual domain adaptation
  challenge}, CoRR abs/1710.06924 (2017).
\newblock \href {http://arxiv.org/abs/1710.06924} {\path{arXiv:1710.06924}}.
\newline\urlprefix\url{http://arxiv.org/abs/1710.06924}

\bibitem{ganin2015unsupervised}
Y.~Ganin, V.~Lempitsky, Unsupervised domain adaptation by backpropagation, in:
  International Conference on Machine Learning, 2015, pp. 1180--1189.

\bibitem{long2017deep}
M.~Long, H.~Zhu, J.~Wang, M.~I. Jordan, Deep transfer learning with joint
  adaptation networks, in: Proceedings of the 34th International Conference on
  Machine Learning-Volume 70, JMLR. org, 2017, pp. 2208--2217.

\bibitem{pei2018multi}
Z.~Pei, Z.~Cao, M.~Long, J.~Wang, Multi-adversarial domain adaptation, in:
  Thirty-Second AAAI Conference on Artificial Intelligence, 2018.

\bibitem{he2016deep}
K.~He, X.~Zhang, S.~Ren, J.~Sun, Deep residual learning for image recognition,
  in: Proceedings of the IEEE conference on computer vision and pattern
  recognition, 2016, pp. 770--778.

\bibitem{russakovsky2015imagenet}
O.~Russakovsky, J.~Deng, H.~Su, J.~Krause, S.~Satheesh, S.~Ma, Z.~Huang,
  A.~Karpathy, A.~Khosla, M.~Bernstein, et~al., Imagenet large scale visual
  recognition challenge, International journal of computer vision 115~(3)
  (2015) 211--252.

\bibitem{DBLP:journals/pami/MiyatoMKI19}
T.~Miyato, S.~Maeda, M.~Koyama, S.~Ishii, Virtual adversarial training: {A}
  regularization method for supervised and semi-supervised learning, {IEEE}
  Trans. Pattern Anal. Mach. Intell. 41~(8) (2019) 1979--1993.

\end{thebibliography}

\end{document}